\documentclass{article}

\PassOptionsToPackage{numbers, compress}{natbib}

\usepackage[preprint]{neurips_2026}

\usepackage[utf8]{inputenc} % allow utf-8 input
\usepackage[T1]{fontenc} % use 8-bit T1 fonts
\usepackage{hyperref} % hyperlinks
\usepackage{url} % simple URL typesetting
\usepackage{booktabs} % professional-quality tables
\usepackage{amsfonts} % blackboard math symbols
\usepackage{nicefrac} % compact symbols for 1/2, etc.
\usepackage{microtype} % microtypography

\usepackage{algorithm}
\usepackage{algpseudocode}
\usepackage{listings}

\usepackage{xcolor} 
\usepackage{graphicx}
\usepackage{adjustbox}
\usepackage{pifont}
\usepackage{amsmath,amssymb}
\usepackage{pgfplots}
\usepackage{comment}
\usepackage{soul}
\usepackage{svg}
\usepackage{wrapfig}
\usepackage{array}
\usepackage{xspace}
\usepackage{subcaption}
\usepackage{mathrsfs}
\usepackage{sidecap}
\usepackage{colortbl} 

\usepackage{bm}
\usepackage{enumitem}
\usepackage{cutwin}
\usepackage{multicol}
\usepackage{multirow}
\definecolor{custompink}{RGB}{255, 20, 147} 
\def\eg{\textit{e.g.}}

\hypersetup{
    colorlinks=true, 
    linkcolor=custompink, 
    filecolor=custompink,
    urlcolor=custompink
}

\title{Towards Consistent Video Geometry Estimation}

\author{
    \small
    Zhu Yu$^{1\dagger}$\thanks{Internship at Tongyi Lab} \quad
    Jingnan Gao$^{3}$\thanks{Equal Contribution} \quad
    Runmin Zhang$^{1}$ \quad
    Lingteng Qiu$^{2}$ \quad
    Zhengyi Zhao$^{2}$ \quad
    Rui Peng$^{2}$\quad
    \\\small
    \textbf{Yichao Yan}$^{3}$ \quad
    \textbf{Kejie Qiu}$^{2}$ \quad 
    \textbf{Siyu Zhu}$^{4}$ \quad
    \textbf{Zilong Dong}$^{2}$ \quad
    \textbf{Si-Yuan Cao}$^{1}$ \quad
    \textbf{Hui-Liang Shen}$^{1}$
    \\[4pt]
    \footnotesize
    $^1$Zhejiang University \quad $^2$Tongyi Lab, Alibaba Group \quad
    $^3$Shanghai Jiao Tong University \quad
    $^4$Fudan University \\
    {\tt\small \href{https://pkqbajng.github.io/ViGeo/}{Project Page: https://pkqbajng.github.io/ViGeo/}}
}

\begin{document}

\maketitle

\vspace{-0.8em}
\begingroup
    \centering
    \captionsetup{type=figure,skip=2pt}
    \includegraphics[width=\textwidth]{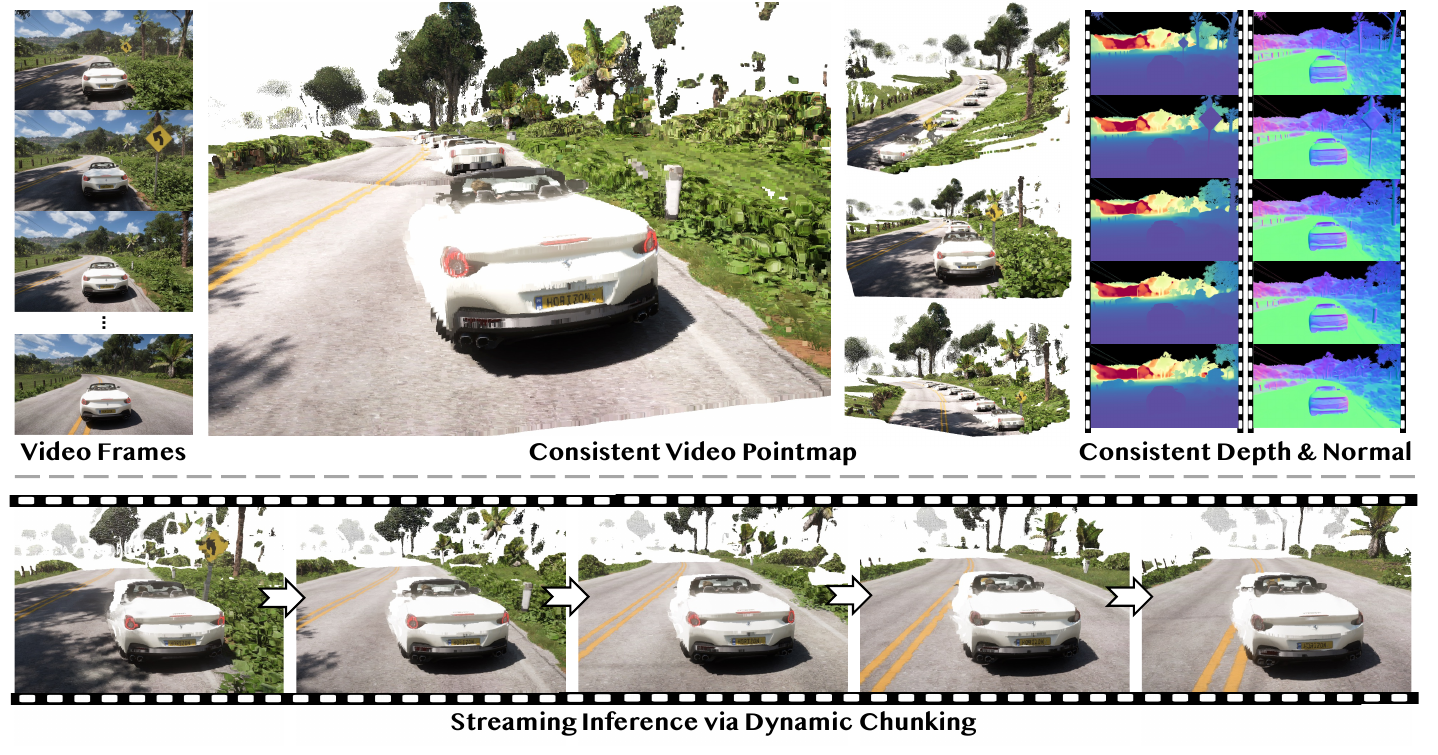}
    \caption{
    \textbf{ViGeo} is a unified feed-forward foundation model for video geometry estimation. 
    It predicts temporally consistent depth, surface normals, and dense point maps from raw video frames. 
    With dynamic chunking attention, the same trained model seamlessly switches between full-sequence reconstruction and streaming inference without retraining.
    }
    
    \label{teaser}
\par
\endgroup
\vspace{0.35em}

\begin{abstract}
\hyphenpenalty=10000
\exhyphenpenalty=10000
\emergencystretch=1em
This work presents \textbf{ViGeo}, a feed-forward foundation model for recovering spatially dense and temporally consistent geometry from video sequences. Built upon a plain transformer architecture without task-specific architectural modifications, ViGeo supports streaming, full-sequence, and long-video inference within a unified model. The key design is \textbf{dynamic chunking attention}, which exposes the model to both bidirectional and causal temporal contexts during training and allows it to adapt its attention pattern at test time without retraining. To improve supervision quality, we further introduce a \textbf{completion-based data refinement framework}. This framework trains a video depth completion teacher that conditions on sparse and noisy annotations and exploits video/multi-view context to produce dense, temporally coherent, and geometrically reliable training targets. Beyond depth and point maps, ViGeo also predicts surface normals within the same framework. Trained solely on public datasets, ViGeo achieves state-of-the-art performance across online, offline, and long-video depth estimation, surface normal estimation, and video point map estimation.
\end{abstract}

\section{Introduction}
\begin{wrapfigure}{r}{0.48\textwidth}
    \centering
    \includegraphics[width=0.48\textwidth]{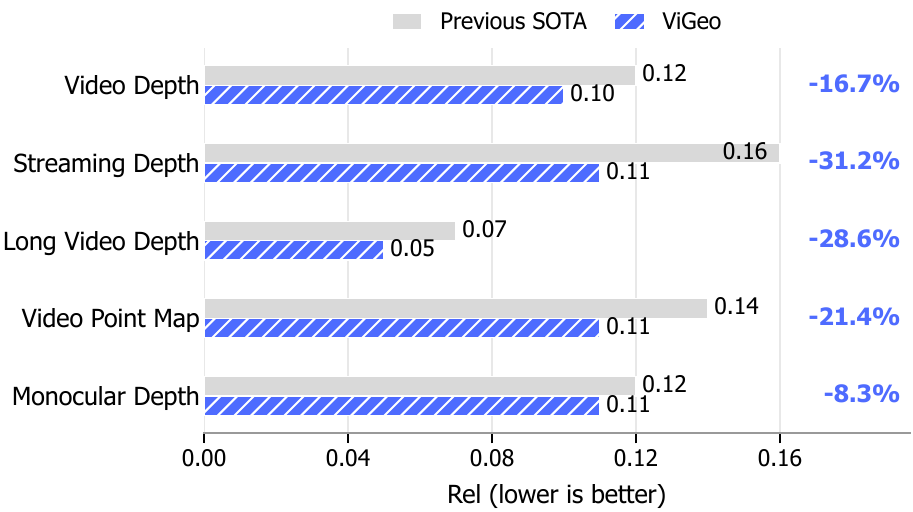}
    \caption{Benchmark comparison with previous state-of-the-art methods.}
    \label{fig:bar}
\end{wrapfigure}
Video geometry estimation is a fundamental problem in computer vision, supporting applications such as robotics~\cite{geovideogen}, augmented reality~\cite{depthar}, autonomous navigation~\cite{depthnav}, and video editing~\cite{depthedit}. These applications require geometry that is both spatially accurate and temporally consistent over long video sequences. Despite recent progress, achieving high-fidelity reconstruction, long-term consistency, and scalable inference within a unified video model remains challenging.

A central limitation of existing video geometry models is their fixed temporal access pattern: offline methods~\cite{vggt,pi3,more,mapanything} rely on future frames for full-sequence reasoning, while online methods~\cite{streamvggt,stream3r,cut3r,spann3r} operate with restricted causal context. As a result, current models cannot adapt their attention behavior to the available video context at inference time. Large-scale training supervision poses another bottleneck: real-captured video geometry datasets are commonly built from LiDAR measurements~\cite{waymo, scannetpp, arkitscenes} or SfM reconstructions~\cite{dl3dv, colmap}, whose sparse, noisy, or scale-ambiguous annotations limit spatial sharpness and temporal consistency.

In this work, we present \textbf{ViGeo}, a feed-forward foundation model for dense and temporally consistent geometry estimation from video sequences. Instead of using separate architectures or training protocols for different inference regimes, ViGeo adopts a plain transformer backbone with \textbf{dynamic chunking attention}. This design exposes the model to both bidirectional and causal temporal contexts during training, allowing it to adapt its attention pattern at inference time without retraining. By changing the chunk partition, ViGeo can operate in full-sequence, streaming, and long-video settings, while remaining compatible with key-value (KV) caching~\cite{infinitevggt} for long-sequence processing.

To improve supervision from real-captured data, we further introduce a \textbf{completion-based data refinement framework} for scalable video geometry learning. Rather than treating raw annotations as reliable ground truth, we view them as imperfect geometric observations that should be completed and rectified. Prior refinement pipelines often rely on monocular depth prediction, followed by either affine alignment to sparse observations~\cite{depth_anything_v1,depth_anything_v2,da3} or reconstruction-based post-processing~\cite{moge2}. In contrast, our framework trains a video depth completion teacher that conditions on sparse and noisy annotations while leveraging temporal and multi-view context to produce dense, temporally coherent, and geometrically reliable training targets. This refinement process can be applied across diverse real-captured datasets, providing a practical data engine for large-scale video geometry supervision.

ViGeo also supports surface normal estimation alongside depth and point map prediction within the same framework. To reflect the practical requirements of video geometry estimation, we evaluate ViGeo across streaming, offline, and long-video depth estimation, as well as surface normal and point map estimation. Trained solely on publicly available datasets, ViGeo achieves state-of-the-art results on most metrics and remains competitive on the rest.

Our contributions are summarized as follows:

\begin{enumerate}
    \item We present \textbf{ViGeo}, a feed-forward foundation model for dense and temporally consistent video geometry estimation. Built upon a plain transformer backbone, ViGeo supports depth, surface normal, and point map estimation within a unified framework.
    \item We introduce \textbf{dynamic chunking attention}, which exposes the model to both bidirectional and causal temporal contexts during training. This design enables a single trained model to adapt to full-sequence, streaming, and long-video inference without retraining, and remains compatible with KV caching for scalable long-sequence processing.
    \item We propose a \textbf{completion-based data refinement framework} that trains a video depth completion teacher to refine sparse and noisy LiDAR/SfM annotations into dense, temporally coherent, and geometrically reliable training targets.
    
    \item We conduct extensive evaluations across multiple datasets and benchmarks, covering streaming, offline, and long-video depth estimation, as well as surface normal and point map estimation. ViGeo achieves state-of-the-art performance and demonstrates strong generalization across diverse video geometry settings.
\end{enumerate}

\section{Related Work}
\label{sec:related_work}

\textbf{Dense monocular geometry estimation.} Early methods~\cite{eigendepth, dorn, vnl_mono, adabins, crfdepth, bae2021estimating, qi2020geonet, pointdc} are generally restricted to in-domain datasets, severely limiting their generalization to unseen environments. MiDaS~\cite{midas, dpt, midasv3} pioneers a paradigm shift: by introducing an affine-invariant objective, it unifies diverse data sources for large-scale joint training, drastically improving zero-shot capabilities for relative depth estimation. Building upon this trajectory, subsequent approaches~\cite{depth_anything_v1, depth_anything_v2, moge, metric3d, metric3dv2, unidepth, unidepthv2, zerodepth} further scale the training corpus, while recent diffusion-based methods~\cite{marigold, stablediffusion, fe2e} successfully harness the strong generative priors of latent diffusion models. More recently, drawing inspiration from multi-task learning, joint depth and surface normal estimation methods~\cite{dens3r, moge2, fe2e, geowizard, lotus, dsine, marigoldv2} have emerged, effectively leveraging the mutual benefits of these complementary geometric representations. Despite significant progress, by processing images in isolation, existing monocular estimators naturally lack multi-view geometric consistency, leading to severe scale ambiguities and temporal flickering across different viewpoints.

\textbf{Dense video geometry estimation.} Moving beyond isolated frames, dense video geometry estimation fundamentally aims to recover temporally coherent and spatially accurate geometry from video sequences. A common approach is to jointly optimize depth across multiple images using classic and learnable dense visual SLAM methods~\cite{drioid, vggsfm}, or to globally align the outputs of single-image estimators~\cite{nvds, luo2020consistent}. Recently, some approaches~\cite{align3r, monst3r} have also demonstrated that DUSt3R~\cite{dust3r} can generalize to videos and dynamic scenes. However, a shared bottleneck across all these paradigms is their heavy reliance on computationally expensive post-optimization. Driven by the rapid advancements in feed-forward foundation models~\cite{vggt, more, dens3r, cut3r, fast3r, pi3, mvdust3r, mast3r, vda, depthcrafter, chronodepth, depthanyvideo, da3}, the trend has recently shifted towards directly predicting consistent geometry from video sequences in a purely feed-forward manner. Building upon the robust priors of Depth Anything~\cite{depth_anything_v2}, Video Depth Anything~\cite{vda} devises an efficient spatiotemporal head and a temporal consistency loss to enforce temporal coherence. Concurrently, DepthCrafter~\cite{depthcrafter} unleashes the potential of latent video diffusion models~\cite{svd} to generate highly consistent, open-world video depth sequences. Furthermore, recent feed-forward 3D reconstruction models~\cite{vggt, pi3, mast3r, more} have demonstrated that temporal coherence can also be effectively achieved via alternating attention mechanisms, while concurrently revealing that multi-task learning paradigms (\eg, joint estimation of depth, point maps, and surface normals) significantly enhance overall geometric representation. However, the majority of these architectures are primarily designed for offline inference, where the full sequence is available, and are not naturally suited for streaming or causal settings. To address sequential input scenarios, several recent works have explored streaming 3D reconstruction with causal or persistent memory designs~\cite{cut3r, spann3r, streamvggt, stream3r, flashdepth}. CUT3R~\cite{cut3r} introduces a continuous 3D perception model with a persistent state, while FlashDepth~\cite{flashdepth} leverages a recurrent network to perform online alignment. More recently, StreamVGGT~\cite{streamvggt} and Stream3R~\cite{stream3r} extend large-scale geometric transformers to streaming settings through causal architectures. Although these methods improve scalability for long sequences and enable online inference, they typically trade off global context and reconstruction quality compared to offline models. As a result, existing approaches still separate offline and streaming reconstruction into different model designs, leaving a critical gap for a unified framework that can flexibly handle both regimes without retraining.

\textbf{Large-scale data training.}
Recently, large-scale data training coupled with advanced network backbones~\cite{dinov2, dpt, vit} has emerged as a powerful paradigm for 3D geometry estimation~\cite{vggt,pi3,more,dens3r,stream3r, streamvggt}. Due to the lack of high-quality labeled 3D datasets, various data engines~\cite{depth_anything_v1, depth_anything_v2, moge2, da3} have been devised. Depth Anything~\cite{depth_anything_v1, depth_anything_v2} scales the training datasets by unleashing the power of large-scale unlabeled data, but such paradigms are fundamentally restricted to relative disparity estimation. To obtain more reliable 3D supervision, MoGe2~\cite{moge2} enhances the annotations of noisy datasets~\cite{waymo} via Poisson reconstruction, while Depth Anything 3~\cite{da3} directly aligns monocular depth maps with sparse measurements. However, these approaches heavily rely on the initial outputs of monocular depth estimation, leaving the final refined annotations inherently bounded by the errors of the underlying monocular estimators. Inspired by the progress of depth completion foundation models~\cite{ldcm, lingbodepth}, we devise a data engine based on multi-view depth completion, fully leveraging the strengths from both images and sparse measurements to generate dense, accurate, and temporally consistent depth annotations for large-scale training.

\section{Method}

\begin{figure}[t] 
    \centering
    \includegraphics[width=\linewidth]{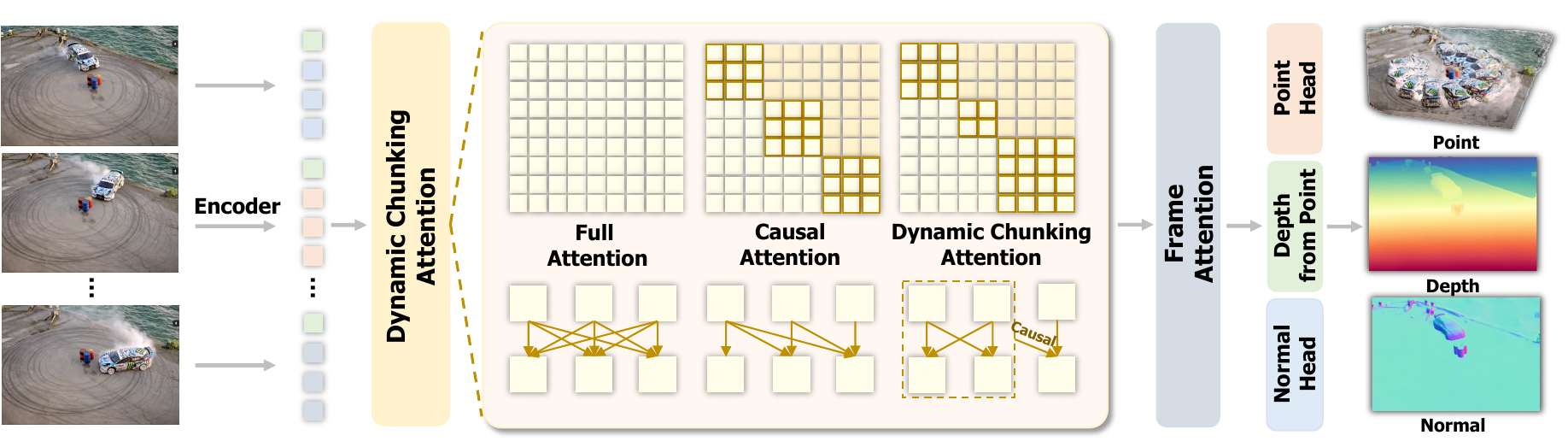} 
    \caption{
    Architecture overview of \textbf{ViGeo}. 
    Built upon a plain Transformer with dynamic chunking attention, ViGeo supports full-sequence, streaming, and long-video inference within a unified model and predicts temporally consistent depth, surface normals, and point maps.}
    \label{fig:architecture}
\end{figure}

This section presents the methodology of \textbf{ViGeo}, a unified feed-forward framework for consistent monocular video geometry estimation. We first describe the overall network architecture in Sec.~\ref{sec:overall_arc}. Sec.~\ref{sec:attn} then introduces dynamic chunking attention, which enables a single trained model to adapt to streaming, full-sequence, and long-video inference without retraining. Next, Sec.~\ref{sec:refine} describes our completion-based data refinement framework, which trains a video depth completion teacher to construct dense and temporally coherent supervision from sparse and noisy real-world annotations. Finally, Sec.~\ref{sec:training} formulates the training objectives used to optimize ViGeo.

\subsection{Overall Architecture}
\label{sec:overall_arc}

Given a video clip of $N$ RGB frames $\{\mathbf{I}_i\}_{i=1}^N$, ViGeo predicts dense geometric quantities for each frame in a fully feed-forward manner. As illustrated in Fig.~\ref{fig:architecture}, ViGeo is built upon a plain ViT-style Transformer. The early layers operate within individual frames to extract dense visual tokens, while the later layers alternate between intra-frame attention and dynamic chunking attention to jointly model spatial details and temporal dependencies. The resulting spatiotemporal features are then decoded into point maps, depth maps, and surface normals. Formally, ViGeo maps the input sequence to a set of per-frame geometric predictions:
\begin{equation}
    \{\mathbf{P}_i, \mathbf{D}_i, \mathbf{N}_i\}_{i=1}^N = f\big(\{\mathbf{I}_i\}_{i=1}^N\big),
\end{equation}
where $\mathbf{P}_i \in \mathbb{R}^{3 \times H \times W}$ denotes the point map, $\mathbf{D}_i \in \mathbb{R}^{H \times W}$ denotes the depth map, and $\mathbf{N}_i \in \mathbb{R}^{3 \times H \times W}$ denotes the surface normal map of frame $i$.

Our architecture follows the recent trend of large feed-forward geometric models, but is designed for a more flexible video inference setting. Instead of committing to either full-sequence attention~\cite{vggt, pi3, da3} or causal attention~\cite{streamvggt, stream3r}, ViGeo employs dynamic chunking attention to bridge these temporal access patterns within a single model. Together with intra-frame attention, this design keeps the backbone simple and generic, while allowing the same trained model to operate under offline, streaming, and long-video inference without architectural modification or retraining.

\subsection{Dynamic Chunking Attention Design}
\label{sec:attn}

Existing video geometry models usually adopt either full-sequence bidirectional attention for offline reconstruction~\cite{vggt, pi3, da3, more} or causal attention for streaming inference~\cite{streamvggt, stream3r}. Instead of fixing the temporal access pattern, we introduce \textbf{dynamic chunking attention}, which allows a single model to adapt its attention behavior through the chunk partition. Given a sequence of frames, tokens attend bidirectionally within the same chunk and causally across different chunks. In other words, attention is full within each chunk and causal across chunks.

Formally, we partition the input sequence into a set of contiguous temporal chunks:
\begin{equation}
    \mathcal{N} = \{\mathcal{N}_1, \mathcal{N}_2, \dots, \mathcal{N}_L\}, \qquad \sum_{l=1}^{L} |\mathcal{N}_l| = N,
\end{equation}
where $|\mathcal{N}_l|$ denotes the number of consecutive frames in the $l$-th chunk, and $N$ is the total sequence length. Let $\mathrm{ch}(i)$ denote the chunk index of frame $i$. We define a frame-level attention mask $\mathcal{M}^{\text{attn}}$, where the entry between query frame $i$ and key frame $j$ is given by:
\begin{equation}
    \mathcal{M}^{\text{attn}}_{i,j} = 
    \begin{cases} 
    1, & \mathrm{ch}(j) \leq \mathrm{ch}(i), \\
    0, & \text{otherwise}.
    \end{cases}
    \label{eq:attn_mask}
\end{equation}
This mask is applied to all visual tokens according to their frame indices. When two frames belong to the same chunk, they can attend to each other bidirectionally. When they belong to different chunks, a frame can only attend to frames from previous chunks.

\begin{table}[t]
\centering
\caption{
    Inference modes induced by dynamic chunking attention. By changing only the temporal chunk partition, the same attention formulation covers full-sequence, streaming, and chunk-based inference while sharing the same model parameters.
}
\begin{tabular}{lccc}
    \toprule
    Mode & Chunk & Intra-Chunk & Inter-Chunk \\
    \midrule
    Full-sequence (Offline) & $[\mathcal{N}]$ & Full attention & -- \\
    Chunk-based & $[\mathcal{N}_1,\dots,\mathcal{N}_L]$ & Full attention & Causal attention \\
    Streaming (Online) & $[1,\dots,1]$ & -- & Causal attention \\
    \bottomrule
\end{tabular}
\label{tab:attn_modes}
\end{table}

As summarized in Table~\ref{tab:attn_modes}, different chunk partitions instantiate different inference modes under the same formulation. When $L=1$, all frames belong to a single chunk, and Eq.~\ref{eq:attn_mask} reduces to full-sequence bidirectional attention for offline inference. When $|\mathcal{N}_l|=1$ for all $l$, each chunk contains one frame and the mask becomes strictly causal, enabling streaming inference. Intermediate chunk sizes induce chunk-based inference, preserving bidirectional context within local temporal groups while maintaining causal access across chunks. During training, we expose the model to multiple chunk configurations, including both bidirectional and causal temporal contexts. At inference time, the same trained model can switch among full-sequence, streaming, and long-video settings by specifying the chunk partition, without modifying the architecture or retraining.

Dynamic chunking attention also supports scalable long-video processing. For long sequences, chunk-based inference is compatible with KV caching~\cite{infinitevggt}, allowing past states to be reused across chunks and helping control memory growth. This formulation also fits practical streaming scenarios, where inputs may arrive in short multi-frame packets.

\subsection{Completion-Based Data Refinement}
\label{sec:refine}

\begin{figure}[t]
    \centering
    \includegraphics[width=\linewidth]{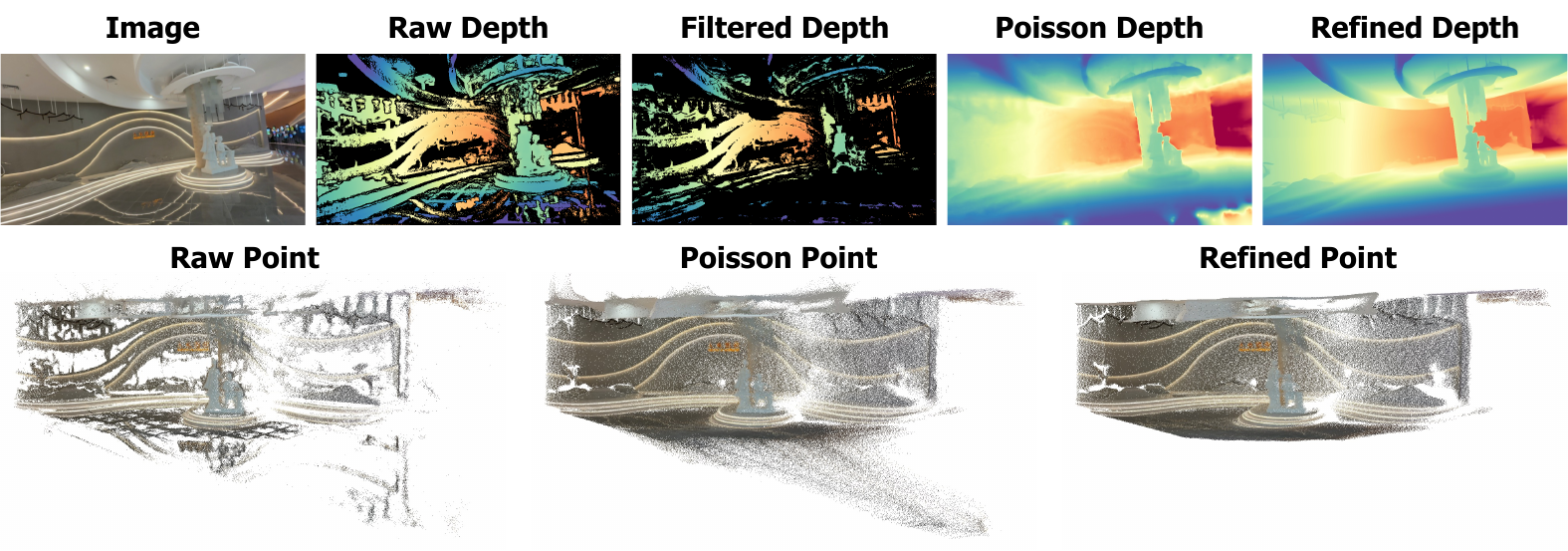}
    \caption{
    Visualization of our completion-based data refinement pipeline.
    Given an RGB video sequence and raw depth maps with missing regions and noisy measurements, we first apply per-frame outlier filtering to obtain reliable sparse observations, and then construct dense but coarse depth priors via Poisson reconstruction.
    A video depth completion teacher refines these priors by leveraging temporal and multi-view context, producing sharp and coherent dense depth labels.
    The bottom row compares the corresponding point clouds, where our refined labels reduce missing regions, flying points, and geometric artifacts.
    }
    \label{fig:refine_pipeline}
\end{figure}

In practice, real-captured depth annotations~\cite{dl3dv, wildrgb, Blendedmvs, arkitscenes} often contain missing regions, outliers, and scale ambiguity. Directly using such measurements as supervision can degrade spatial fidelity and temporal consistency. Prior refinement pipelines~\cite{moge2, da3, depth_anything_v1, depth_anything_v2} often rely on monocular depth predictions, followed by alignment to sparse observations or reconstruction-based post-processing. In contrast, we formulate real-data supervision refinement as a video depth completion problem. Our pipeline treats raw annotations as imperfect geometric observations and trains a video depth completion teacher, which is then used to produce dense, temporally coherent, and geometrically reliable pseudo-labels. As shown in Fig.~\ref{fig:refine_pipeline}, the pipeline consists of two stages: per-frame outlier filtering and multi-frame video depth completion.

\noindent\textbf{Outlier Filtering.}
We first filter unreliable raw measurements before depth completion. Given a raw depth map $\mathbf{D}^{\text{raw}}$, we use the local spherical alignment criterion from MoGe-2~\cite{moge2} to identify inconsistent observations, yielding a valid mask $\mathcal{M}^{\text{valid}}$. The filtered sparse depth is obtained as $\tilde{\mathbf{D}} = \mathbf{D}^{\text{raw}} \odot \mathcal{M}^{\text{valid}}$, where $\odot$ denotes the Hadamard product.

\noindent\textbf{Video Depth Completion Teacher.}
Given the filtered sparse depth sequence, we use the trained video depth completion teacher to generate dense pseudo-labels. To provide a dense geometric condition for the teacher, we first convert the filtered sparse depth $\tilde{\mathbf{D}}$ into a coarse dense prior $\mathbf{D}^{\text{prior}}$ using Poisson reconstruction~\cite{ldcm}. Following LDCM~\cite{ldcm}, the prior is obtained by aligning its log-gradient field with that of an initial monocular relative depth prediction $\mathbf{D}^{\text{mono}}$, while preserving the reliable sparse measurements in $\tilde{\mathbf{D}}$:
\begin{equation}
    \min_{\mathbf{D}^{\text{prior}}} 
    \sum_{p} \left\|\nabla \log \mathbf{D}^{\text{prior}}_p - \nabla \log(\mathbf{D}^{\text{mono}} + \gamma)_p\right\|^2 
    + \lambda \sum_{p\in\mathcal{M}^{\text{valid}}} 
    \left(\mathbf{D}^{\text{prior}}_p - \tilde{\mathbf{D}}_p\right)^2,
\end{equation}
where $\gamma$ is a shift factor derived from affine alignment~\cite{ldcm}. Unlike prior pipelines that directly use the monocular prediction or its aligned reconstruction as the final supervision, this prior only serves as a dense geometric condition for the video completion teacher.

Given the full sequence of RGB images $\{\mathbf{I}_i\}_{i=1}^N$ and the corresponding dense priors $\{\mathbf{D}^{\text{prior}}_i\}_{i=1}^N$, we apply median-based log normalization to handle scale-ambiguous data~\cite{dl3dv, Blendedmvs}, such as SfM reconstructions. Specifically, the dense priors are normalized as $\log(\mathbf{D}^{\text{prior}}_i / m)$, where $m$ is the median depth value computed over valid sparse measurements across the temporal sequence. Following LingBot-Depth~\cite{lingbodepth}, RGB images and normalized depth priors are separately embedded as patch tokens with spatial and modality-specific positional encodings.

The teacher adopts a ViT-style architecture similar to ViGeo, extending this RGB-prior completion formulation from single images to video sequences. Its deeper layers aggregate intra-frame and cross-frame context to complete the coarse priors into temporally coherent dense depth predictions. The predicted depth is restored to the original scale by multiplying it with the sequence median $m$, yielding the final dense pseudo-labels:
\begin{equation}
    \{\hat{\mathbf{D}}^{\text{pseudo}}_i\}_{i=1}^N =
    \mathcal{F}^{\text{teacher}}
    \left(
    \{\mathbf{I}_i\}_{i=1}^N,
    \left\{\log(\mathbf{D}^{\text{prior}}_i / m)\right\}_{i=1}^N
    \right) \cdot m.
\end{equation}
These dense pseudo-labels $\{\hat{\mathbf{D}}^{\text{pseudo}}_i\}_{i=1}^N$ replace the raw measurements as supervision for training ViGeo on real-captured data.

Fig.~\ref{fig:refine_pipeline} illustrates the effect of each refinement stage. Raw measurements contain missing regions and outliers. Poisson reconstruction densifies the depth but may introduce flying points and geometric artifacts. The video depth completion teacher further refines these priors, producing denser and more coherent point clouds that better align with image structures. More qualitative examples are provided in Sec.~\ref{sec:qualitative_results}.

\subsection{Training Objectives}
\label{sec:training}
We train ViGeo end-to-end with a multi-task geometry loss:
\begin{equation}
    \mathcal{L} =
    \mathcal{L}_{\text{points}}
    + \lambda_{\text{points\_normal}}\mathcal{L}_{\text{points\_normal}}
    + \lambda_{\text{normal}}\mathcal{L}_{\text{normal}} .
\end{equation}
Since depth is directly obtained from the predicted point map, we supervise the 3D point map as the primary geometric representation. The point map loss $\mathcal{L}_{\text{points}}$ penalizes the $L_1$ distance between the predicted and ground-truth point maps:
\begin{equation}
    \mathcal{L}_{\text{points}} =
    \sum_{i=1}^{N} \sum_{p \in \mathcal{M}^{\text{valid}}}
    \frac{1}{\hat{\mathbf{D}}_{i,p}}
    \left\| s \mathbf{P}_{i,p} - \hat{\mathbf{P}}_{i,p} \right\|_1 ,
\end{equation}
where $\mathbf{P}_{i,p}$ and $\hat{\mathbf{P}}_{i,p}$ denote the predicted and ground-truth point maps at pixel $p$ of frame $i$, respectively, and $\hat{\mathbf{D}}_{i,p}$ denotes the ground-truth depth. To handle scale ambiguity, the scale factor $s$ is estimated by minimizing:
\begin{equation}
    s =
    \arg\min_{s'}
    \sum_{i=1}^{N} \sum_{p \in \mathcal{M}^{\text{valid}}}
    \frac{1}{\hat{\mathbf{D}}_{i,p}}
    \left\| s' \mathbf{P}_{i,p} - \hat{\mathbf{P}}_{i,p} \right\|_1 ,
\end{equation}
which is efficiently solved using the ROE solver~\cite{moge}.

In addition to point-wise supervision, we impose surface geometry constraints with two normal-related losses. The direct normal loss $\mathcal{L}_{\text{normal}}$ supervises the predicted normal map $\mathbf{N}_{i,p}$ using angular distance:
\begin{equation}
    \mathcal{L}_{\text{normal}} =
    \sum_{i=1}^{N} \sum_{p \in \mathcal{M}^{\text{valid}}}
    \arccos \left(
    \frac{\mathbf{N}_{i,p}^{\top}\hat{\mathbf{N}}_{i,p}}
    {\|\mathbf{N}_{i,p}\| \|\hat{\mathbf{N}}_{i,p}\|}
    \right),
    \label{eq:normal}
\end{equation}
where $\hat{\mathbf{N}}_{i,p}$ denotes the ground-truth surface normal. We further introduce a geometry-derived normal loss $\mathcal{L}_{\text{points\_normal}}$, which follows the same angular formulation as Eq.~\ref{eq:normal} but replaces the explicit normal prediction with the normal analytically computed from the predicted point map $\mathbf{P}_{i,p}$. This loss encourages the predicted 3D structure to preserve locally coherent surface geometry. The video depth completion teacher is optimized with the same loss formulation as LDCM~\cite{ldcm}.

\subsection{Implementation Details}
\label{sec:implementation_details}
ViGeo adopts ViT-G~\cite{vit} as the backbone. Following recent feed-forward geometry models~\cite{da3}, one-third of the attention layers are configured with dynamic chunking attention. The backbone features are passed to a 5-layer transformer decoder that applies self-attention within each frame, followed by separate convolutional heads~\cite{moge} for geometric prediction.

Training is conducted in two stages. In the first stage, ViGeo is trained for 50K iterations with a fixed pixel budget of 112,896. In the second stage, it is fine-tuned for 200K iterations with variable resolutions, where the pixel budget is randomly sampled between 112,896 and 268,324. Across both stages, the batch size varies from 2 to 24 samples, and the aspect ratio is randomly sampled from $[0.5, 2.0]$. The backbone is initialized from the pretrained DA3 weights~\cite{da3}. After the two-stage training, we freeze the preceding network modules and independently optimize the confidence head following Pi3~\cite{pi3}.

We use the AdamW optimizer with a cosine learning rate schedule and linear warmup. The peak learning rates are set to $5 \times 10^{-5}$ and $1 \times 10^{-5}$ for the first and second stages, respectively, and the backbone learning rate is scaled by 0.1. We apply standard data augmentations, including random cropping, color jittering, Gaussian blur, JPEG compression artifacts, and perspective-aware cropping. The first stage is trained on 16 NVIDIA H20 GPUs, while the second stage uses 48 NVIDIA H20 GPUs. The complete training process takes approximately 20 days.

\noindent\textbf{Training data.}
We collect 23 open-source RGB-D datasets to train ViGeo, comprising 17 synthetic and 6 real-world datasets. An overview of the training datasets is provided in Table~\ref{tab:training_datasets}, spanning five distinct domains: indoor, outdoor, driving, object, and in-the-wild scenarios. The combined training set covers a diverse range of environments, from controlled synthetic spaces to complex real-world captures utilizing LiDAR and 3D reconstruction techniques. The number of scenes and RGB-D pairs in each dataset may slightly differ from the originally released versions, as we manually filtered the data to exclude invalid frames and ensure high-quality training samples. Specifically, surface normal supervision is restricted to only high-quality synthetic datasets.

\begin{table}[h]
	\centering
	\caption{An overview of the training datasets.}
	\resizebox{\linewidth}{!}{
		\begin{tabular}{cccc}
			\toprule
			Dataset & Domain & \#Scenes & Type \\
			\midrule
			Hypersim~\cite{hypersim} & Indoor & 741 & Synthetic \\
			LightWheelOcc~\cite{lightwheelocc} & Outdoor/Driving & 1020 & Synthetic \\
			TartanAir~\cite{tartanair} & In-the-wild & 412 & Synthetic \\
            TartanGround~\cite{tartanground} & In-the-wild & 8100 & Synthetic \\
			GTA-SfM~\cite{gtasfm} & Outdoor/In-the-wild & 213 & Synthetic \\
			PointOdyssey~\cite{pointodyssey} & Indoor & 33 & Synthetic \\
			BEDLAM~\cite{bedlam} & Indoor & 7309 & Synthetic \\
			Dynamic Replica~\cite{dynamicreplica} & Indoor & 1046 & Synthetic \\
			MatrixCity~\cite{matrixcity} & Outdoor/Driving & 2581 & Synthetic \\
			MVS-Synth~\cite{mvssynth} & Outdoor/Driving & 120 & Synthetic \\
			OmniWorld~\cite{omniworld} & In-the-wild & 11424 & Synthetic \\
			Synthia~\cite{synthia} & Outdoor/Driving & 284 & Synthetic \\
			OmniObject3D~\cite{omniobject3d} & Object & 4794 & Synthetic \\
			Aria Synthetic Environments~\cite{ase} & Indoor & 22946 & Synthetic \\
			Spring~\cite{spring} & In-the-wild & 35 & Synthetic \\
			TransPhy3D~\cite{transphy3d} & Transparent/Reflective & 11148 & Synthetic \\
			CarlaOcc~\cite{carlaocc} & Outdoor/Driving & 591 & Synthetic \\
			WildRGB~\cite{wildrgb} & Indoor/Object & 2058 & LiDAR \\
			Waymo~\cite{waymo} & Outdoor/Driving & 3790 & LiDAR \\
			ARKitScenes~\cite{arkitscenes} & Indoor & 4307 & LiDAR \\
			ScanNet++~\cite{scannetpp} & Indoor & 1811 & LiDAR \\
			DL3DV~\cite{dl3dv} & In-the-wild & 6379 & COLMAP \\
			BlendedMVS~\cite{Blendedmvs} & In-the-wild & 503 & 3D Reconstruction \\
			\bottomrule
		\end{tabular}
	}
	\label{tab:training_datasets}
\end{table}

\section{Experiments}
\subsection{Evaluation Protocol}
\label{sec:evaluation_protocol}

\subsubsection{Evaluation Datasets}
\label{sec:evaluation_datasets}

ViGeo is evaluated across multiple benchmark datasets and inference settings. Unless otherwise specified, input images are resized to satisfy each model's resolution requirements, and predictions are resized back to the original resolution before metric computation. We prepare the evaluation datasets as follows:

\begin{itemize}
    \item \textbf{Sintel}~\cite{sintel}: We use all sequences from the training split for evaluation. Each sequence contains 21 to 50 frames with an original resolution of $1024 \times 436$. Sintel is used for monocular depth estimation, video depth estimation, video point map estimation, surface normal estimation, and camera pose estimation. For depth evaluation, the maximum depth is set to 70 meters. For pose estimation, we use the clean image pass and evaluate the full sequences with the provided camera trajectories.

    \item \textbf{Bonn}~\cite{bonn}: Following Pi3~\cite{pi3}, we evaluate on five Bonn sequences with an original resolution of $640 \times 480$. Bonn is used for monocular depth estimation, video depth estimation, long video depth estimation, and video point map estimation. For standard evaluation, we sample 110 frames from each sequence. For long video depth estimation, we sample 400 consecutive frames starting from the first frame of each sequence.

    \item \textbf{KITTI}~\cite{kitti}: Following Pi3~\cite{pi3}, we evaluate on 13 KITTI sequences with an original resolution of $1242 \times 375$. KITTI is used for monocular depth estimation, video depth estimation, long video depth estimation, and video point map estimation. For standard evaluation, we sample 110 frames from each sequence. For long video depth estimation, we sample 300 consecutive frames from each sequence.

    \item \textbf{HAMMER}~\cite{hammer}: We evaluate on 11 HAMMER sequences with an original resolution of $1088 \times 832$. HAMMER is used for long video depth estimation and surface normal estimation. For long video depth estimation, we sample 300 consecutive frames from each sequence. For surface normal estimation, we sample 110 frames with a stride of 2.

    \item \textbf{NYUv2}~\cite{nyuv2}: We use NYUv2 for monocular surface normal estimation and evaluate on the official test split of 654 images with an original resolution of $640 \times 480$. Following the standard evaluation protocol, images are cropped to $565 \times 427$ before metric computation.

    \item \textbf{7-Scenes}~\cite{sevenscenes}: We use 7-Scenes for 3D reconstruction evaluation. The dataset contains RGB-D sequences from seven indoor scenes with camera poses. Following the reconstruction benchmark protocol, we evaluate on the official test sequences and sample frames with a stride of 200. The raw depth maps are projected into the RGB camera frame before evaluation.

    \item \textbf{Neural RGB-D}~\cite{neuralrgbd}: We use Neural RGB-D (NRGBD) for 3D reconstruction evaluation. The dataset contains nine indoor RGB-D sequences with camera poses. We evaluate on all scenes and sample frames with a stride of 500 following the sparse reconstruction protocol.
\end{itemize}

\subsubsection{Evaluation Metrics}
\label{sec:evaluation_metrics}

We report evaluation metrics for depth estimation, video point map estimation, surface normal estimation, camera pose estimation, and 3D reconstruction. Pixel-wise metrics are computed over valid pixels. We denote the set of valid frame-pixel pairs as $\mathcal{V}$. For scale-ambiguous dense predictions, we first align the prediction with a scalar scale factor $s$ following the protocol in the main paper.

\noindent\textbf{Depth metrics.}
For depth estimation, we report the absolute relative error $\mathrm{Rel}$ and threshold accuracy $\delta_1$. Let $\mathbf{D}_{i,p}$ and $\hat{\mathbf{D}}_{i,p}$ denote the predicted and ground-truth depths at pixel $p$ of frame $i$, respectively. The metrics are defined as:
\begin{equation}
    \mathrm{Rel} =
    \frac{1}{|\mathcal{V}|}
    \sum_{(i,p) \in \mathcal{V}}
    \frac{
    \left|s\mathbf{D}_{i,p} - \hat{\mathbf{D}}_{i,p}\right|
    }{
    \hat{\mathbf{D}}_{i,p}
    },
\end{equation}
\begin{equation}
    \delta_1 =
    \frac{1}{|\mathcal{V}|}
    \sum_{(i,p) \in \mathcal{V}}
    \mathbf{1}
    \left[
    \max
    \left(
    \frac{s\mathbf{D}_{i,p}}{\hat{\mathbf{D}}_{i,p}},
    \frac{\hat{\mathbf{D}}_{i,p}}{s\mathbf{D}_{i,p}}
    \right)
    < 1.25
    \right].
\end{equation}

\noindent\textbf{Point map metrics.}
For video point map estimation, we report the point-wise relative error $\mathrm{Rel}^{p}$ and threshold accuracy $\delta^p_{0.25}$. Let $\mathbf{P}_{i,p}$ and $\hat{\mathbf{P}}_{i,p}$ denote the predicted and ground-truth point maps at pixel $p$ of frame $i$, respectively. The metrics are defined as:
\begin{equation}
    \mathrm{Rel}^{p} =
    \frac{1}{|\mathcal{V}|}
    \sum_{(i,p) \in \mathcal{V}}
    \frac{
    \left\|s\mathbf{P}_{i,p} - \hat{\mathbf{P}}_{i,p}\right\|_2
    }{
    \left\|\hat{\mathbf{P}}_{i,p}\right\|_2
    },
\end{equation}
\begin{equation}
    \delta^p_{0.25} =
    \frac{1}{|\mathcal{V}|}
    \sum_{(i,p) \in \mathcal{V}}
    \mathbf{1}
    \left[
    \frac{
    \left\|s\mathbf{P}_{i,p} - \hat{\mathbf{P}}_{i,p}\right\|_2
    }{
    \left\|\hat{\mathbf{P}}_{i,p}\right\|_2
    }
    < 0.25
    \right].
\end{equation}

\noindent\textbf{Surface normal metrics.}
For surface normal estimation, we compute the angular error between the predicted normal $\mathbf{N}_{i,p}$ and ground-truth normal $\hat{\mathbf{N}}_{i,p}$:
\begin{equation}
    \theta_{i,p} =
    \arccos
    \left(
    \frac{
    \mathbf{N}_{i,p}^{\top}\hat{\mathbf{N}}_{i,p}
    }{
    \|\mathbf{N}_{i,p}\|_2
    \|\hat{\mathbf{N}}_{i,p}\|_2
    }
    \right).
\end{equation}
We report the mean and median angular errors, denoted as $\mathrm{Mean}$ and $\mathrm{Med}$, as well as $\delta_{11.25^\circ}$, the percentage of pixels whose angular error is below $11.25^\circ$:
\begin{equation}
    \delta_{11.25^\circ} =
    \frac{1}{|\mathcal{V}|}
    \sum_{(i,p) \in \mathcal{V}}
    \mathbf{1}
    \left[
    \theta_{i,p} < 11.25^\circ
    \right].
\end{equation}

\noindent\textbf{Camera pose metrics.}
For camera pose estimation, let $\mathbf{T}_i \in SE(3)$ and $\hat{\mathbf{T}}_i \in SE(3)$ denote the predicted and ground-truth camera-to-world poses at frame $i$, respectively. Since monocular pose predictions can be scale-ambiguous, we first align the predicted trajectory to the ground-truth trajectory with a similarity transform, yielding $\tilde{\mathbf{T}}_i$. We report the absolute trajectory error (ATE) as the root mean squared translation error:
\begin{equation}
    \mathrm{ATE} =
    \sqrt{
    \frac{1}{N}
    \sum_{i=1}^{N}
    \left\|
    \mathrm{trans}\left(\tilde{\mathbf{T}}_i\right)
    -
    \mathrm{trans}\left(\hat{\mathbf{T}}_i\right)
    \right\|_2^2
    }.
\end{equation}
We also report relative pose error (RPE) over frame interval $\Delta=1$. Let
\begin{equation}
    \mathbf{E}_i =
    \left(
    \hat{\mathbf{T}}_i^{-1}\hat{\mathbf{T}}_{i+\Delta}
    \right)^{-1}
    \left(
    \tilde{\mathbf{T}}_i^{-1}\tilde{\mathbf{T}}_{i+\Delta}
    \right)
\end{equation}
denote the relative pose residual. The translational and rotational RPEs are:
\begin{equation}
    \mathrm{RPE}_{\mathrm{trans}} =
    \sqrt{
    \frac{1}{N-\Delta}
    \sum_{i=1}^{N-\Delta}
    \left\|
    \mathrm{trans}\left(\mathbf{E}_i\right)
    \right\|_2^2
    },
\end{equation}
\begin{equation}
    \mathrm{RPE}_{\mathrm{rot}} =
    \sqrt{
    \frac{1}{N-\Delta}
    \sum_{i=1}^{N-\Delta}
    \angle\left(\mathrm{rot}\left(\mathbf{E}_i\right)\right)^2
    },
\end{equation}
where $\angle(\cdot)$ denotes the rotation angle in degrees.

\noindent\textbf{3D reconstruction metrics.}
For 3D reconstruction, we convert predicted and ground-truth depth/point maps into point clouds. The predicted point cloud is first aligned to the ground truth using a similarity transform followed by point-to-point ICP. Let $\mathcal{P}$ and $\hat{\mathcal{P}}$ denote the aligned predicted and ground-truth point clouds, respectively, with corresponding surface normals $\mathbf{n}_{\mathbf{p}}$ and $\hat{\mathbf{n}}_{\hat{\mathbf{p}}}$. We define the nearest-neighbor operators
\begin{equation}
    \pi_{\hat{\mathcal{P}}}(\mathbf{p})
    =
    \arg\min_{\hat{\mathbf{p}} \in \hat{\mathcal{P}}}
    \left\|\mathbf{p} - \hat{\mathbf{p}}\right\|_2,
    \quad
    \pi_{\mathcal{P}}(\hat{\mathbf{p}})
    =
    \arg\min_{\mathbf{p} \in \mathcal{P}}
    \left\|\hat{\mathbf{p}} - \mathbf{p}\right\|_2.
\end{equation}
Accuracy and completeness measure nearest-neighbor distances from prediction to ground truth and from ground truth to prediction:
\begin{equation}
    d_{\mathrm{acc}}(\mathbf{p}) =
    \left\|
    \mathbf{p} - \pi_{\hat{\mathcal{P}}}(\mathbf{p})
    \right\|_2,
    \quad
    d_{\mathrm{comp}}(\hat{\mathbf{p}}) =
    \left\|
    \hat{\mathbf{p}} - \pi_{\mathcal{P}}(\hat{\mathbf{p}})
    \right\|_2.
\end{equation}
We report both mean and median statistics of $d_{\mathrm{acc}}$ and $d_{\mathrm{comp}}$. Normal consistency is computed from nearest-neighbor normal agreement in both directions:
\begin{equation}
    \mathrm{NC} =
    \frac{1}{2}
    \left(
    \mathrm{stat}_{\mathbf{p}\in\mathcal{P}}
    \left[
    \left|
    \mathbf{n}_{\mathbf{p}}^\top
    \hat{\mathbf{n}}_{\pi_{\hat{\mathcal{P}}}(\mathbf{p})}
    \right|
    \right]
    +
    \mathrm{stat}_{\hat{\mathbf{p}}\in\hat{\mathcal{P}}}
    \left[
    \left|
    \hat{\mathbf{n}}_{\hat{\mathbf{p}}}^\top
    \mathbf{n}_{\pi_{\mathcal{P}}(\hat{\mathbf{p}})}
    \right|
    \right]
    \right),
\end{equation}
where $\mathrm{stat}$ is instantiated as either mean or median.

\subsection{Main Results}

\subsubsection{Video Depth Estimation}
\label{sec:video_depth_estimation}

\begin{table}[t]
    \centering
    \caption{\textbf{Quantitative results for scale-invariant video depth estimation} on the Sintel~\cite{sintel}, Bonn~\cite{bonn}, and KITTI~\cite{kitti} datasets. Baselines are grouped into offline and online settings, and evaluated via absolute relative error (Rel) and threshold accuracy ($\delta_{1}$). The best and second-best results in each category are highlighted in \textbf{bold} and \underline{underlined}, respectively.}
    
    \setlength{\tabcolsep}{4pt} 
    \renewcommand{\arraystretch}{1.1} 

    \begin{tabular*}{\columnwidth}{@{\extracolsep{\fill}} l cccccc}
        \toprule
        \multirow{3}{*}{\textbf{Method}} &
        \multicolumn{2}{c}{\textbf{Sintel}} &
        \multicolumn{2}{c}{\textbf{Bonn}} &
        \multicolumn{2}{c}{\textbf{KITTI}} \\
        \cmidrule(lr){2-3} \cmidrule(lr){4-5} \cmidrule(lr){6-7}
        & Rel $\downarrow$ & $\delta_{1}$ $\uparrow$ & Rel $\downarrow$ & $\delta_{1}$ $\uparrow$ & Rel $\downarrow$ & $\delta_{1}$ $\uparrow$ \\
        \midrule
        \rowcolor[HTML]{F2F2F2} \multicolumn{7}{l}{\textit{\textbf{Offline}}} \\
        VDA~\cite{vda} & 0.855 & 0.340 & 0.301 & 0.465 & 0.272 & 0.535 \\
        DepthCrafter~\cite{depthcrafter} & 0.469 & 0.455 & 0.074 & 0.968 & 0.282 & 0.435 \\
        GeometryCrafter~\cite{geometrycrafter} & 0.305 & \underline{0.675} & 0.057 & 0.969 & 0.057 & 0.981 \\
        VGGT~\cite{vggt}  & 0.373 & 0.634 & 0.057 & 0.966 & 0.062 & 0.969 \\
        Pi3~\cite{pi3}  & \underline{0.281} & 0.663 & \underline{0.049} & \underline{0.975} & \underline{0.038} & \textbf{0.985} \\
        DA3~\cite{da3}  & 0.284 & 0.666 & 0.050 & 0.972 & 0.058 & 0.976 \\
        \midrule
        ViGeo (offline) & \textbf{0.229} & \textbf{0.724} & \textbf{0.046} & \textbf{0.976} & \textbf{0.036} & \underline{0.983} \\
        \midrule
        \rowcolor[HTML]{F2F2F2} \multicolumn{7}{l}{\textit{\textbf{Online}}} \\
        StreamVGGT~\cite{streamvggt} & 0.407 & 0.614 & \underline{0.059} & 0.971 & 0.173 & 0.721 \\
        Stream3R~\cite{stream3r} & \underline{0.319} & \underline{0.675} & 0.068 & 0.956 & \underline{0.088} & \underline{0.938} \\
        FlashDepth~\cite{flashdepth} & 0.361 & 0.562 & 0.066 & 0.974 & 0.064 & 0.971 \\
        \midrule
        ViGeo (online) & \textbf{0.256} & \textbf{0.683} & \textbf{0.050} & \textbf{0.974} & \textbf{0.049} & \textbf{0.977} \\
        \bottomrule
    \end{tabular*}
    \label{tab:scale_invariant_video_depth}
\end{table}

We evaluate ViGeo for video depth estimation on the Sintel~\cite{sintel}, Bonn~\cite{bonn}, and KITTI~\cite{kitti} datasets, with quantitative results summarized in Table~\ref{tab:scale_invariant_video_depth}. To comprehensively assess our model, we compare it against a diverse set of baselines that encompass both offline and online paradigms. These include prominent video depth estimators (VideoDepthAnything~\cite{vda}, DepthCrafter~\cite{depthcrafter}, FlashDepth~\cite{flashdepth}), as well as state-of-the-art 3D reconstruction and depth foundation models (VGGT~\cite{vggt}, Pi3~\cite{pi3}, Depth Anything 3~\cite{da3}, CUT3R~\cite{cut3r}, StreamVGGT~\cite{streamvggt}, Stream3R~\cite{stream3r}). Following standard practice, all predicted depth sequences are aligned to the ground truth using a single scale factor per sequence. We report the absolute relative error (Abs Rel) and threshold accuracy ($\delta_1$). The results demonstrate the robust effectiveness of our method across various scenarios, maintaining consistently high performance in both settings. Notably, even when restricted to the online streaming mode, ViGeo surpasses several existing offline methods.

\subsubsection{Long Video Depth Estimation}
\label{sec:long_video_depth_estimation}
\begin{table}[t]
    \centering
    \caption{\textbf{Quantitative results for scale-invariant long video depth estimation} on the Bonn~\cite{bonn}, KITTI~\cite{kitti}, and HAMMER~\cite{hammer} datasets. We report the absolute relative error (Rel) and threshold accuracy ($\delta_{1}$). The best and second-best results are highlighted in \textbf{bold} and \underline{underlined}, respectively.}
    
    \renewcommand{\arraystretch}{1.1}
    \setlength{\tabcolsep}{0pt} 

    \begin{tabular*}{\columnwidth}{@{\extracolsep{\fill}} l cccccc}
        \toprule
        {\multirow{3}{*}{\textbf{Method}}} &
        \multicolumn{2}{c}{\textbf{Bonn (400)}} &
        \multicolumn{2}{c}{\textbf{KITTI (300)}} &
        \multicolumn{2}{c}{\textbf{HAMMER (300)}} \\
        \cmidrule(lr){2-3} \cmidrule(lr){4-5} \cmidrule(lr){6-7}
        & 
        Rel $\downarrow$ & $\delta_{1}$ $\uparrow$ &
        Rel $\downarrow$ & $\delta_{1}$ $\uparrow$ &
        Rel $\downarrow$ & $\delta_{1}$ $\uparrow$ \\
        \midrule
        VGGT~\cite{vggt}  & \textcolor{gray}{OOM} & \textcolor{gray}{OOM} &  \textcolor{gray}{OOM} & \textcolor{gray}{OOM} & \textcolor{gray}{OOM} & \textcolor{gray}{OOM} \\
        StreamVGGT~\cite{streamvggt} & \textcolor{gray}{OOM} & \textcolor{gray}{OOM} &  \textcolor{gray}{OOM} & \textcolor{gray}{OOM} & \textcolor{gray}{OOM} & \textcolor{gray}{OOM} \\
        Stream3R~\cite{stream3r} & \textcolor{gray}{OOM} & \textcolor{gray}{OOM} &  \textcolor{gray}{OOM} & \textcolor{gray}{OOM} & \textcolor{gray}{OOM} & \textcolor{gray}{OOM} \\
        \midrule
        VDA~\cite{vda} & 0.291 & 0.531 & 0.274 & 0.465 & 0.278 & 0.482 \\
        DepthCrafter~\cite{depthcrafter} & 0.132 & 0.891 & 0.288 & 0.451 & 0.490 & 0.302\\
        GeometryCrafter~\cite{geometrycrafter} & 0.101 & 0.934 & \underline{0.070} & \underline{0.970} & \underline{0.054} & \underline{0.983}\\
        FlashDepth~\cite{flashdepth} &  0.084 & 0.943 & 0.100 & 0.919 & 0.064 & 0.971 \\
        InfiniteVGGT~\cite{infinitevggt} & \underline{0.069} & \underline{0.965} & 0.206 & 0.659 & 0.093 & 0.921 \\
        \midrule
        ViGeo & \textbf{0.059} & \textbf{0.967} & \textbf{0.052} & \textbf{0.977} & \textbf{0.019} & \textbf{0.996} \\
        \bottomrule
    \end{tabular*}
    \label{tab:scale_invariant_long_video_depth}
\end{table}

To evaluate the long video depth estimation capability of ViGeo, we benchmark our model on Bonn~\cite{bonn}, KITTI~\cite{kitti}, and HAMMER~\cite{hammer} with extended sequences of 300--400 frames. As shown in Table~\ref{tab:scale_invariant_long_video_depth}, several strong baselines, including VGGT~\cite{vggt} and its streaming variants~\cite{streamvggt, stream3r}, encounter out-of-memory (OOM) errors and fail to process these long sequences. In contrast, ViGeo handles long videos through dynamic chunking attention, which adapts the temporal access pattern at inference time and remains compatible with key-value (KV) caching~\cite{infinitevggt}. Quantitative results show that ViGeo achieves state-of-the-art performance across all three datasets, outperforming both offline and online baselines. Its consistently low relative error and high threshold accuracy ($\delta_1$) indicate strong temporal stability and robustness on extended video sequences.

\subsubsection{Video Point Map Estimation}
\begin{table}[t]
    \centering
    \caption{\textbf{Quantitative results for scale-invariant video point map estimation} on the Sintel~\cite{sintel}, Bonn~\cite{bonn}, and KITTI~\cite{kitti} datasets. Baselines are grouped into offline and online settings, and evaluated via point-wise relative error ($\text{Rel}^{p}$) and threshold accuracy ($\delta^{p}_{0.25}$) computed over the 3D point coordinates. The best and second-best results in each category are highlighted in \textbf{bold} and \underline{underlined}, respectively.}
    
    \renewcommand{\arraystretch}{1.1}
    \setlength{\tabcolsep}{0pt} 

    \begin{tabular*}{\columnwidth}{@{\extracolsep{\fill}} l cccccc}
        \toprule
        {\multirow{3}{*}{\textbf{Method}}} &
        \multicolumn{2}{c}{\textbf{Sintel}} &
        \multicolumn{2}{c}{\textbf{Bonn}} &
        \multicolumn{2}{c}{\textbf{KITTI}} \\
        \cmidrule(lr){2-3} \cmidrule(lr){4-5} \cmidrule(lr){6-7}
        & 
        $\text{Rel}^{p}$ $\downarrow$ & $\delta^{p}_{0.25}$ $\uparrow$ &
        $\text{Rel}^{p}$ $\downarrow$ & $\delta^{p}_{0.25}$ $\uparrow$ &
        $\text{Rel}^{p}$ $\downarrow$ & $\delta^{p}_{0.25}$ $\uparrow$ \\
        \midrule
        \rowcolor[HTML]{F2F2F2} \multicolumn{7}{l}{\textit{\textbf{Offline}}} \\
        GeometryCrafter~\cite{geometrycrafter} & 0.280 & 0.642 & 0.070 & 0.975 & 0.084 & 0.984\\
        VGGT~\cite{vggt} & 0.268 & \underline{0.670} & 0.065 & \underline{0.976} & 0.092 & 0.967 \\
        Pi3~\cite{pi3} & \underline{0.257} & 0.661 & 0.097 & 0.821 & \underline{0.073} & \textbf{0.992} \\
        DA3~\cite{da3} & 0.285 & 0.589 & \underline{0.059} & 0.975 & 0.148 & 0.935 \\
        \midrule
        ViGeo (offline) & \textbf{0.219} & \textbf{0.751} & \textbf{0.052} & \textbf{0.978} & \textbf{0.050} & \underline{0.986} \\
        \midrule
        \rowcolor[HTML]{F2F2F2} \multicolumn{7}{l}{\textit{\textbf{Online}}} \\
        StreamVGGT~\cite{streamvggt} & 0.300 & \underline{0.654} & \underline{0.070} & \textbf{0.977} & 0.180 & 0.768 \\
        Stream3R~\cite{stream3r} & \underline{0.266} & 0.646 & 0.090 & \underline{0.976} & \underline{0.102} & \underline{0.953} \\
        \midrule
        ViGeo (online) & \textbf{0.234} & \textbf{0.714} & \textbf{0.059} & \textbf{0.977} & \textbf{0.063} & \textbf{0.981} \\
        \bottomrule
    \end{tabular*}
    \label{tab:scale_invariant_video_pointmap}
\end{table}

To assess the ability of ViGeo to recover dense 3D geometry, we evaluate video point map estimation on the same datasets used for video depth estimation. Since point maps represent per-pixel 3D geometry, we report the point-wise relative error ($\mathrm{Rel}^{p}$) and threshold accuracy ($\delta^p_{0.25}$) over 3D point coordinates. As shown in Table~\ref{tab:scale_invariant_video_pointmap}, ViGeo achieves strong geometric accuracy and consistently outperforms existing baselines across all evaluated datasets. These results demonstrate that ViGeo recovers accurate dense 3D geometry in video sequences.

\subsubsection{Surface Normal Estimation}
\begin{table}[htbp]
    \centering
    \caption{\textbf{Quantitative results for surface normal estimation} on the Sintel~\cite{sintel}, HAMMER~\cite{hammer}, and NYUv2~\cite{nyuv2} datasets. We report the mean and median angular errors, as well as the threshold accuracy ($\delta_{11.25^\circ} \uparrow$). The best and second-best results are highlighted in \textbf{bold} and \underline{underlined}, respectively.}
    
    \renewcommand{\arraystretch}{1.2}
    \setlength{\tabcolsep}{0pt} 

    \begin{tabular*}{\columnwidth}{@{\extracolsep{\fill}} l ccccccccc}
        \toprule
        \multirow{2}{*}{\textbf{Method}} &
        \multicolumn{3}{c}{\textbf{Sintel}} &
        \multicolumn{3}{c}{\textbf{HAMMER}} &
        \multicolumn{3}{c}{\textbf{NYUv2}} \\
        \cmidrule(lr){2-4} \cmidrule(lr){5-7} \cmidrule(lr){8-10}
        & Mean $\downarrow$ & Med $\downarrow$ & $\delta_{11.25^\circ}$ $\uparrow$ 
        & Mean $\downarrow$ & Med $\downarrow$ & $\delta_{11.25^\circ}$ $\uparrow$ 
        & Mean $\downarrow$ & Med $\downarrow$ & $\delta_{11.25^\circ}$ $\uparrow$ \\
        \midrule
        DSINE~\cite{dsine} & 40.69 & 35.52 & 21.22 & 12.90 & 8.36 & 68.21 & 16.50 & 9.17 & 58.75 \\
        NormalCrafter~\cite{normalcrafter} & \textbf{35.39} & \textbf{28.57} & \underline{24.34} & 11.11 & 7.83 & 70.05 & \underline{15.87} & \underline{8.70} & \underline{60.41} \\
        StableNormal~\cite{stablenormal} & 40.74 & 33.31 & 19.04 & \underline{9.54} & \underline{6.82} & \textbf{80.62} & 17.55 & 10.26 & 55.90 \\
        Lotus~\cite{lotus} & 37.37 & 29.89 & 24.03 & 12.06 & 9.47 & 61.53 & 17.25 & 9.43 & 58.00 \\
        \midrule
        ViGeo & \underline{36.93} & \underline{28.89} & \textbf{26.05} & \textbf{9.32} & \textbf{6.16} & \underline{78.29} & \textbf{15.11} & \textbf{8.51} & \textbf{61.30} \\
        % Ours (online) & & & & & & & & & \\
        \bottomrule
    \end{tabular*}
    \label{tab:surface_normal_estimation}
\end{table}

For surface normal estimation, we evaluate ViGeo on the Sintel~\cite{sintel} and HAMMER~\cite{hammer} video datasets, as well as the NYUv2~\cite{nyuv2} image benchmark. We compare against the image-based methods DSINE~\cite{dsine}, StableNormal~\cite{stablenormal}, and Lotus~\cite{lotus}, as well as the video-based normal estimator NormalCrafter~\cite{normalcrafter}. Following standard protocols, we report the mean and median angular errors (Mean $\downarrow$, Med $\downarrow$) and the threshold accuracy within $11.25^\circ$ ($\delta_{11.25^\circ} \uparrow$). As shown in Table~\ref{tab:surface_normal_estimation}, ViGeo achieves the best mean and median angular errors on HAMMER and the highest threshold accuracy on Sintel, indicating strong normal estimation quality on video sequences. On NYUv2, ViGeo achieves the best results across all metrics, despite being designed as a unified video geometry model. These results show that ViGeo supports reliable surface normal estimation alongside depth and point map prediction.

\subsubsection{Monocular Depth Estimation}
\begin{table}[t]
    \centering
    \caption{\textbf{Quantitative results for monocular depth estimation} on the Sintel~\cite{sintel}, Bonn~\cite{bonn}, and KITTI~\cite{kitti} datasets. We report the absolute relative error (Rel) and threshold accuracy ($\delta_{1}$). The best and second-best results are highlighted in \textbf{bold} and \underline{underlined}, respectively.}

    \renewcommand{\arraystretch}{1.1}
    \setlength{\tabcolsep}{0pt} 

    \begin{tabular*}{\columnwidth}{@{\extracolsep{\fill}} l cccccc}
        \toprule
        {\multirow{3}{*}{\textbf{Method}}} &
        \multicolumn{2}{c}{\textbf{Sintel}} &
        \multicolumn{2}{c}{\textbf{Bonn}} &
        \multicolumn{2}{c}{\textbf{KITTI}} \\
        \cmidrule(lr){2-3} \cmidrule(lr){4-5} \cmidrule(lr){6-7}
        & 
        Rel $\downarrow$ & $\delta_{1}$ $\uparrow$ &
        Rel $\downarrow$ & $\delta_{1}$ $\uparrow$ &
        Rel $\downarrow$ & $\delta_{1}$ $\uparrow$ \\
        \midrule
        VDA~\cite{vda} & 0.529 & 0.407 & 0.349 & 0.460 & 0.256 & 0.580 \\
        DepthCrafter~\cite{depthcrafter} & 0.291 & 0.619 & 0.084 & 0.928 & 0.192 & 0.678 \\
        %GeometryCrafter~\cite{geometrycrafter} & \textbf{0.213} & \underline{0.729} & 0.052 & 0.969 & 0.056 & \underline{0.974} \\
        VGGT~\cite{vggt} & \underline{0.254} & 0.707 & \underline{0.047} & \underline{0.974} &  0.077 &  0.929\\
        Pi3~\cite{pi3} & 0.272 & \textbf{0.724} & \textbf{0.044} & \textbf{0.975} & \underline{0.055} & \textbf{0.977} \\
        DA3~\cite{da3}  & 0.281 & 0.673 & 0.053 & 0.963 & 0.073 & 0.954 \\
        FlashDepth~\cite{flashdepth} & 0.288 & 0.664 & 0.061 & 0.967 & 0.084 & 0.939 \\
        \midrule
        ViGeo & \textbf{0.240} & \underline{0.720} & 0.049 & 0.973 & \textbf{0.054} & \underline{0.971} \\
        \bottomrule
    \end{tabular*}
    \label{tab:monocular_depth}
\end{table}
For monocular depth estimation, we evaluate ViGeo on Sintel~\cite{sintel}, Bonn~\cite{bonn}, and KITTI~\cite{kitti}. This setting assesses the single-frame geometry estimation capability of ViGeo by applying the model independently to each image, without using temporal context. We compare against representative monocular and feed-forward geometry baselines under the same evaluation protocol. Following standard monocular depth evaluation, we apply affine-invariant alignment, i.e., scale and shift alignment, between the predicted and ground-truth depth for each frame. As shown in Table~\ref{tab:monocular_depth}, ViGeo achieves the best performance on Sintel and remains competitive on Bonn and KITTI. These results indicate that, beyond its temporal modeling capability, ViGeo also learns strong single-frame geometric priors and produces reliable depth estimates under the monocular setting.

\subsubsection{3D Reconstruction}
\begin{table}[t]
    \centering
    \caption{\textbf{Quantitative 3D reconstruction results} on the 7-Scenes and NRGBD datasets. We report accuracy (Acc), completeness (Comp), and normal consistency (NC), with mean and median statistics. Baselines are grouped into offline and online settings. The best and second-best results in each category are highlighted in \textbf{bold} and \underline{underlined}, respectively.}
    
    \renewcommand{\arraystretch}{1.1}
    \setlength{\tabcolsep}{2pt}
    \resizebox{\columnwidth}{!}{%
    \begin{tabular}{l cccccc cccccc}
        \toprule
        \multirow{3}{*}{\textbf{Method}} &
        \multicolumn{6}{c}{\textbf{7-Scenes}} &
        \multicolumn{6}{c}{\textbf{NRGBD}} \\
        \cmidrule(lr){2-7} \cmidrule(lr){8-13}
        & \multicolumn{2}{c}{Acc $\downarrow$} &
        \multicolumn{2}{c}{Comp $\downarrow$} &
        \multicolumn{2}{c}{NC $\uparrow$} &
        \multicolumn{2}{c}{Acc $\downarrow$} &
        \multicolumn{2}{c}{Comp $\downarrow$} &
        \multicolumn{2}{c}{NC $\uparrow$} \\
        \cmidrule(lr){2-3} \cmidrule(lr){4-5} \cmidrule(lr){6-7}
        \cmidrule(lr){8-9} \cmidrule(lr){10-11} \cmidrule(lr){12-13}
        & Mean & Med. & Mean & Med. & Mean & Med. &
        Mean & Med. & Mean & Med. & Mean & Med. \\
        \midrule
        \rowcolor[HTML]{F2F2F2} \multicolumn{13}{l}{\textit{\textbf{Offline}}} \\
        VGGT~\cite{vggt} & \textbf{0.043} & \textbf{0.025} & \textbf{0.063} & \textbf{0.038} & \textbf{0.763} & \textbf{0.869} & 0.047 & 0.025 & 0.061 & 0.033 & 0.908 & 0.991 \\
        Pi3~\cite{pi3} & 0.050 & \underline{0.029} & 0.075 & \underline{0.048} & 0.740 & 0.839 & \textbf{0.027} & \textbf{0.015} & \underline{0.030} & \textbf{0.014} & \underline{0.917} & \underline{0.994} \\
        DA3~\cite{da3} & 0.057 & 0.039 & 0.082 & 0.059 & 0.738 & 0.839 & \underline{0.029} & \underline{0.016} & \textbf{0.029} & \textbf{0.014} & \textbf{0.927} & \textbf{0.996} \\
        \midrule
        ViGeo (offline) & \underline{0.049} & \underline{0.029} & \underline{0.070} & \textbf{0.038} & \underline{0.741} & \underline{0.848} & 0.039 & 0.024 & 0.037 & \underline{0.019} & 0.893 & 0.990 \\
        \midrule
        \rowcolor[HTML]{F2F2F2} \multicolumn{13}{l}{\textit{\textbf{Online}}} \\
        StreamVGGT~\cite{streamvggt} & \underline{0.103} & \underline{0.059} & \textbf{0.094} & \textbf{0.056} & \textbf{0.721} & \textbf{0.825} & \underline{0.071} & \underline{0.040} & \underline{0.059} & \underline{0.030} & \underline{0.881} & \textbf{0.992} \\
        Stream3R~\cite{stream3r} & 0.104 & 0.060 & 0.130 & 0.088 & 0.708 & \underline{0.813} & 0.092 & 0.043 & 0.079 & 0.046 & 0.871 & 0.987 \\
        \midrule
        ViGeo (online) & \textbf{0.082} & \textbf{0.057} & \underline{0.111} & \underline{0.086} & \underline{0.709} & 0.804 & \textbf{0.048} & \textbf{0.030} & \textbf{0.040} & \textbf{0.021} & \textbf{0.887} & \underline{0.990} \\
        \bottomrule
    \end{tabular}%
    }
    \label{tab:reconstruction}
\end{table}

We further evaluate 3D reconstruction on the 7-Scenes~\cite{sevenscenes} and NRGBD~\cite{neuralrgbd} datasets. Following prior feed-forward reconstruction benchmarks, we report reconstruction accuracy, completeness, and normal consistency. As shown in Table~\ref{tab:reconstruction}, ViGeo achieves competitive offline reconstruction quality on 7-Scenes and remains close to strong feed-forward baselines on NRGBD. In the online setting, ViGeo improves accuracy and completeness over streaming baselines on both datasets, demonstrating that the same model can recover spatially coherent 3D geometry under causal inference.

\subsubsection{Long-Video 3D Reconstruction}
\begin{table}[t]
    \centering
    \caption{\textbf{Quantitative results for long-video 3D reconstruction} on the 7-Scenes~\cite{sevenscenes} and NRGBD~\cite{neuralrgbd} datasets with 300 input frames. We report accuracy (Acc), completeness (Comp), and normal consistency (NC); each cell shows mean/median statistics. Long-sequence ICP and nearest-neighbor metrics are computed using the same deterministic 500K valid-point sampling for all methods. The best and second-best results are highlighted in \textbf{bold} and \underline{underlined}, respectively.}
    
    \renewcommand{\arraystretch}{1.1}
    \setlength{\tabcolsep}{2pt}
    \resizebox{\columnwidth}{!}{%
    \begin{tabular}{l ccc ccc}
        \toprule
        \multirow{3}{*}{\textbf{Method}} &
        \multicolumn{3}{c}{\textbf{7-Scenes (300)}} &
        \multicolumn{3}{c}{\textbf{NRGBD (300)}} \\
        \cmidrule(lr){2-4} \cmidrule(lr){5-7}
        & Acc $\downarrow$ & Comp $\downarrow$ & NC $\uparrow$ &
        Acc $\downarrow$ & Comp $\downarrow$ & NC $\uparrow$ \\
        \midrule
        VGGT~\cite{vggt} & \textcolor{gray}{OOM} & \textcolor{gray}{OOM} & \textcolor{gray}{OOM} & \textcolor{gray}{OOM} & \textcolor{gray}{OOM} & \textcolor{gray}{OOM} \\
        StreamVGGT~\cite{streamvggt} & \textcolor{gray}{OOM} & \textcolor{gray}{OOM} & \textcolor{gray}{OOM} & \textcolor{gray}{OOM} & \textcolor{gray}{OOM} & \textcolor{gray}{OOM} \\
        Stream3R~\cite{stream3r} & \textcolor{gray}{OOM} & \textcolor{gray}{OOM} & \textcolor{gray}{OOM} & \textcolor{gray}{OOM} & \textcolor{gray}{OOM} & \textcolor{gray}{OOM} \\
        \midrule
        InfiniteVGGT~\cite{infinitevggt} &
        \underline{0.048 / 0.024} & \underline{0.025 / 0.012} & \underline{0.647 / 0.732} &
        \underline{0.055 / 0.035} & \underline{0.025 / 0.012} & \underline{0.715 / 0.855} \\
        \midrule
        ViGeo (chunk) &
        \textbf{0.017 / 0.009} & \textbf{0.014 / 0.007} & \textbf{0.670 / 0.767} &
        \textbf{0.023 / 0.013} & \textbf{0.012 / 0.006} & \textbf{0.761 / 0.894} \\
        \bottomrule
    \end{tabular}%
    }
    \label{tab:long_video_reconstruction}
\end{table}
We further evaluate long-video 3D reconstruction using 300 input frames on 7-Scenes~\cite{sevenscenes} and NRGBD~\cite{neuralrgbd}. We adopt the same reconstruction metrics as in the standard 3D reconstruction benchmark and exclude sequences with fewer than 300 sampled frames. As shown in Table~\ref{tab:long_video_reconstruction}, InfiniteVGGT~\cite{infinitevggt} supports long-context inference through KV-cache-based acceleration, while ViGeo processes long sequences using chunk-based inference with a chunk size of 16. ViGeo consistently improves reconstruction accuracy, completeness, and normal consistency over InfiniteVGGT on both datasets, demonstrating its scalability to long-sequence dense 3D reconstruction.

\subsubsection{Camera Pose Estimation}
We also evaluate camera pose estimation on Sintel. We compare ViGeo with offline feed-forward baselines and streaming reconstruction models using ATE and relative pose errors. As shown in Table~\ref{tab:pose_estimation}, ViGeo achieves competitive offline trajectory accuracy and the tied-best translation RPE among offline methods. Under online inference, ViGeo outperforms streaming baselines across all pose metrics, indicating that dynamic chunking attention provides useful temporal context for camera motion estimation in streaming scenarios.

\subsection{Analysis and Ablation}
\begin{table}[t]
    \centering
    \caption{\textbf{Quantitative results for camera pose estimation} on the Sintel dataset. We report absolute trajectory error (ATE), relative pose error for translation (RPE Trans), and relative pose error for rotation (RPE Rot). Baselines are grouped into offline and online settings. The best and second-best results in each category are highlighted in \textbf{bold} and \underline{underlined}, respectively.}
    
    \renewcommand{\arraystretch}{1.1}
    \setlength{\tabcolsep}{8pt}

    \begin{tabular}{l ccc}
        \toprule
        \multirow{2}{*}{\textbf{Method}} &
        \multicolumn{3}{c}{\textbf{Sintel}} \\
        \cmidrule(lr){2-4}
        & ATE $\downarrow$ & RPE Trans $\downarrow$ & RPE Rot $\downarrow$ \\
        \midrule
        \rowcolor[HTML]{F2F2F2} \multicolumn{4}{l}{\textit{\textbf{Offline}}} \\
        VGGT~\cite{vggt} & 0.420 & 0.196 & 0.396 \\
        Pi3~\cite{pi3} & 0.225 & \underline{0.145} & \textbf{0.189} \\
        DA3~\cite{da3} & \textbf{0.128} & \textbf{0.078} & \underline{0.238} \\
        \midrule
        ViGeo (offline) & \underline{0.184} & \textbf{0.078} & 0.406 \\
        \midrule
        \rowcolor[HTML]{F2F2F2} \multicolumn{4}{l}{\textit{\textbf{Online}}} \\
        StreamVGGT~\cite{streamvggt} & \underline{0.822} & \underline{0.285} & \underline{0.562} \\
        Stream3R~\cite{stream3r} & 0.981 & 0.311 & 0.574 \\
        \midrule
        ViGeo (online) & \textbf{0.382} & \textbf{0.118} & \textbf{0.450} \\
        \bottomrule
    \end{tabular}
    \label{tab:pose_estimation}
\end{table}

\subsubsection{Ablation Study}
\begin{table}[t]
    \centering
    \caption{\textbf{Ablation study on model architecture.} We evaluate the versatility of ViGeo by comparing variants trained with different attention strategies. All models are evaluated under both offline (full-sequence) and online (streaming) inference modes on regular video benchmarks.}
    \label{tab:dynamic_chunking_attention}
    
    \setlength{\tabcolsep}{0pt} 
    \renewcommand{\arraystretch}{1.2} 

    \begin{tabular*}{\columnwidth}{@{\extracolsep{\fill}} l cccccc}
        \toprule
        \multirow{2}{*}{\textbf{Training Scheme}} & \multicolumn{2}{c}{\textbf{Sintel}} & \multicolumn{2}{c}{\textbf{Bonn}} & \multicolumn{2}{c}{\textbf{KITTI}} \\
        \cmidrule(lr){2-3} \cmidrule(lr){4-5} \cmidrule(lr){6-7}
        & Rel $\downarrow$ & $\delta_{1}$ $\uparrow$ & Rel $\downarrow$ & $\delta_{1}$ $\uparrow$ & Rel $\downarrow$ & $\delta_{1}$ $\uparrow$ \\
        \midrule
        \rowcolor[HTML]{F2F2F2} \multicolumn{7}{l}{\textit{\textbf{Offline}}} \\
        Full Attention & \textbf{0.291} & \textbf{0.626} & \textbf{0.052} & \underline{0.971} & \textbf{0.047} & \textbf{0.978} \\
        Causal Attention & 0.325 & 0.592 & 0.057 & 0.969 & 0.058 & 0.974 \\
        Dynamic Chunking (Ours) & \underline{0.301} & \underline{0.609} & \underline{0.054} & \textbf{0.974} & \underline{0.048} & \underline{0.977} \\
        \midrule
        \rowcolor[HTML]{F2F2F2} \multicolumn{7}{l}{\textit{\textbf{Online}}} \\
        Full Attention & 0.365 & 0.549 & 0.072 & 0.955 & 0.169 & 0.672 \\
        Causal Attention & \underline{0.333} & \underline{0.558} & \textbf{0.066} & \textbf{0.963} & \underline{0.057} & \textbf{0.972} \\
        Dynamic Chunking (Ours) & \textbf{0.332} & \textbf{0.568} & \underline{0.067} & \underline{0.961} & \textbf{0.055} & \textbf{0.972} \\
        \bottomrule
    \end{tabular*}
\end{table}

\noindent\textbf{Dynamic Chunking Attention.}
To validate the effectiveness of dynamic chunking attention, we train three ViGeo variants with different temporal attention schemes: full-sequence attention, causal attention, and the proposed dynamic chunking attention. All variants are evaluated under both offline full-sequence inference and online streaming inference. As shown in Table~\ref{tab:dynamic_chunking_attention}, full-attention training performs well in the offline setting but degrades noticeably when evaluated online, due to the mismatch between bidirectional training and causal inference. This degradation is particularly pronounced on KITTI, where fast camera motion and large scene depth ranges make the model more sensitive to changes in temporal context. In contrast, causal-attention training is better aligned with online inference, but it cannot fully exploit bidirectional context under offline evaluation. Dynamic chunking exposes the model to both bidirectional and causal temporal contexts during training, leading to a better overall trade-off across inference regimes. It achieves competitive offline performance while maintaining strong online results, allowing the same trained model to operate under both settings without retraining.

\begin{table}[t]
    \centering
    \caption{\textbf{Ablation study on data refinement framework.} We compare the performance of models trained using raw sensor measurements versus our refined pseudo-labels. All variants are evaluated on the Sintel, Bonn, and KITTI benchmarks.}
    \label{tab:data_refinement}
    
    \setlength{\tabcolsep}{6pt} 
    \renewcommand{\arraystretch}{1.2} 

    \begin{tabular*}{\columnwidth}{@{\extracolsep{\fill}} l cccccc}
        \toprule
        \multirow{2}{*}{\textbf{Training Supervision}} & \multicolumn{2}{c}{\textbf{Sintel}} & \multicolumn{2}{c}{\textbf{Bonn}} & \multicolumn{2}{c}{\textbf{KITTI}} \\
        \cmidrule(lr){2-3} \cmidrule(lr){4-5} \cmidrule(lr){6-7}
        & Rel $\downarrow$ & $\delta_{1}$ $\uparrow$ & Rel $\downarrow$ & $\delta_{1}$ $\uparrow$ & Rel $\downarrow$ & $\delta_{1}$ $\uparrow$ \\
        \midrule
        Raw Measurements & 0.301 & 0.609 & \textbf{0.054} & 0.974 & 0.048 & \textbf{0.977} \\
        \textbf{With Refined Labels (Ours)} & \textbf{0.294} & \textbf{0.619} & 0.055 & \textbf{0.975} & \textbf{0.047} & \textbf{0.977} \\
        \bottomrule
    \end{tabular*}
\end{table}

\noindent\textbf{Completion-Based Data Refinement.}
To evaluate the impact of our data refinement framework, we compare ViGeo trained with raw measurements against the variant supervised by refined labels. As shown in Table~\ref{tab:data_refinement}, refined supervision improves most metrics under the same architecture and training protocol, indicating that higher-quality supervision benefits video geometry learning. This gain comes from the video depth completion teacher, which converts sparse and noisy observations into dense and temporally coherent training targets.

\begin{figure}[t]
    \centering
    \includegraphics[width=\linewidth]{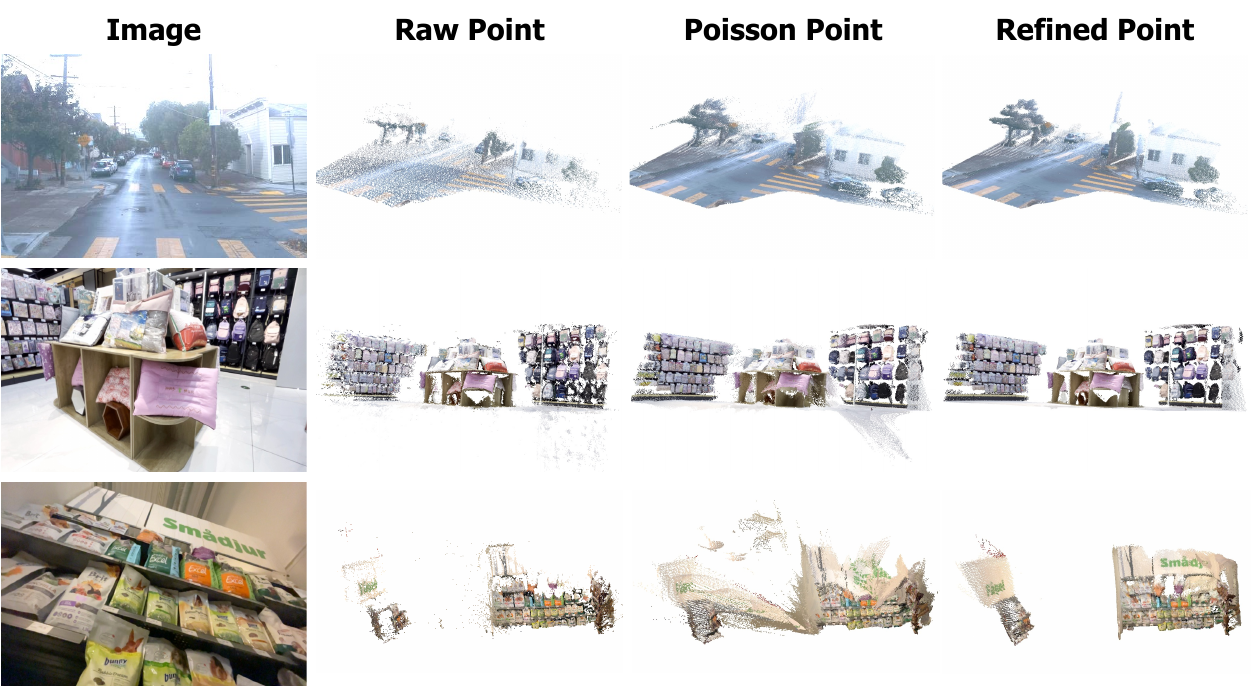}
    \caption{Qualitative ablation of our data refinement pipeline. Our full pipeline effectively resolves the severe missing regions in raw points and the flying points introduced by Poisson reconstruction, yielding clean and structurally coherent 3D geometries.}
    \label{fig:refine_ablation}
\end{figure}

We further provide qualitative comparisons in Fig.~\ref{fig:refine_ablation}. Raw sensor measurements often contain missing regions and outliers. Poisson reconstruction~\cite{ldcm, moge2} densifies the observations but may introduce flying points and geometric distortions. In contrast, our full refinement pipeline produces cleaner and more complete point clouds with better structural coherence.

\subsubsection{Inference Strategy for Long-Video Processing}
\begin{table}[htbp]
    \centering
    \caption{\textbf{Ablation study on inference schemes for long video depth estimation.} All evaluations are performed on sequences of 300$\sim$400 frames. Peak VRAM is measured on the Bonn~\cite{bonn} dataset with a per-image token count of 1369. ``OOM'' indicates out-of-memory on a 96 GB GPU.}
    \label{tab:inference_ablation}
    
    \setlength{\tabcolsep}{0pt} 
    \renewcommand{\arraystretch}{1.2} 

    \begin{tabular*}{\columnwidth}{@{\extracolsep{\fill}} l cccccc c}
        \toprule
        \multirow{2}{*}{\textbf{Scheme}} & \multicolumn{2}{c}{\textbf{Bonn}} & \multicolumn{2}{c}{\textbf{KITTI}} & \multicolumn{2}{c}{\textbf{HAMMER}} & \textbf{VRAM} \\
        \cmidrule(lr){2-3} \cmidrule(lr){4-5} \cmidrule(lr){6-7}
        & Rel $\downarrow$ & $\delta_{1}$ $\uparrow$ & Rel $\downarrow$ & $\delta_{1}$ $\uparrow$ & Rel $\downarrow$ & $\delta_{1}$ $\uparrow$ & \textbf{(GB)} $\downarrow$ \\
        \midrule
        Full Sequence & \textcolor{gray}{OOM} & \textcolor{gray}{OOM} & \textcolor{gray}{OOM} & \textcolor{gray}{OOM} & \textcolor{gray}{OOM} & \textcolor{gray}{OOM} & $>$ 96.0 \\
        \midrule
        Chunked ($C=16$) & 0.059 & 0.967 & 0.052 & 0.977 & 0.019 & 0.996 & \textbf{23.03} \\
        Chunked ($C=32$) & 0.059 & 0.967 & 0.057 & 0.973 & 0.022 & 0.997 & 23.91 \\
        Chunked ($C=48$) & 0.059 & 0.968 & 0.055 & 0.975 & 0.023 & 0.997 & 26.86 \\
        Chunked ($C=64$) & 0.059 & 0.967 & 0.057 & 0.974 & 0.023 & 0.997 & 29.22 \\
        \bottomrule
    \end{tabular*}
\end{table}

Benefiting from our proposed dynamic chunking attention design, our model is able to process arbitrarily long video sequences via a KV-cache mechanism~\cite{infinitevggt}. We ablate different inference schemes by varying the chunk length $C \in \{16, 32, 48, 64\}$. As shown in Table~\ref{tab:inference_ablation}, standard full-sequence inference (i.e., without chunking) scales quadratically in memory complexity, inevitably leading to out-of-memory (OOM) errors on long video sequences ($\sim$400 frames). In contrast, our chunking mechanism strictly bounds the memory footprint by limiting the maximum KV-cache size. Although the peak VRAM increases slightly with larger $C$ due to the overhead of intermediate activations, the depth estimation accuracy remains highly consistent across different chunk sizes. Consequently, we adopt $C=16$ as our default setting to minimize the instantaneous memory peak while maintaining robust temporal consistency for long-range video depth estimation.

\subsubsection{Inference Efficiency}
\begin{table}[htbp]
    \centering
    \small
    \caption{
    \textbf{Efficiency comparison.}
    We report the maximum number of images processed on a 96 GB NVIDIA H20 GPU, model parameters, and average running speed per image.
    Running speed is measured using 16 images at a resolution of $518 \times 518$.
    }
    \label{tab:efficiency}
    \begin{tabular}{lccc}
        \toprule
        Model & Max \# Images & Params. & Speed \\
        \midrule
        VGGT~\cite{vggt} & 200--250 & 1.26B & 10.96 FPS \\
        ViGeo (ours) & 700--800 & 1.26B & 10.00 FPS \\
        \bottomrule
    \end{tabular}
\end{table}

Table~\ref{tab:efficiency} compares the full-attention inference capacity and running speed of ViGeo and VGGT~\cite{vggt} under the same 96 GB NVIDIA H20 GPU setting. Although both models have the same parameter count, ViGeo supports a larger full-attention input capacity, processing 700--800 images compared with 200--250 images for VGGT. Meanwhile, ViGeo maintains a comparable running speed, achieving 10.00 FPS versus 10.96 FPS on 16 images at $518 \times 518$ resolution. It is worth noting that this table reports the maximum number of images under full-attention inference. For long video inference, ViGeo can process arbitrarily long sequences through dynamic chunking attention, which is compatible with KV caching.

\subsection{Qualitative Results}
\label{sec:qualitative_results}

\subsubsection{3D Reconstruction}
\begin{figure}[htbp]
    \centering
    \includegraphics[width=\linewidth]{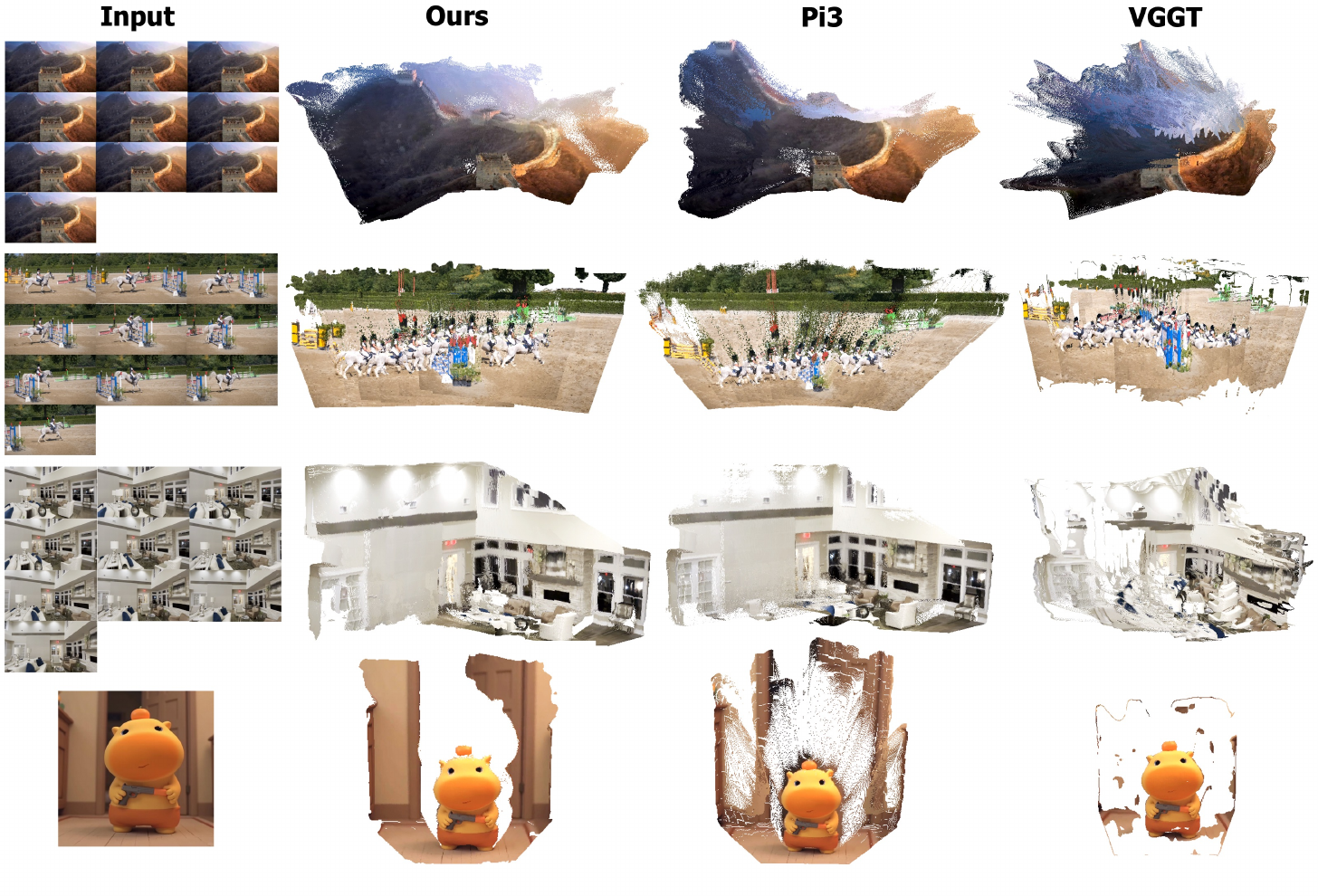}
    \caption{Qualitative results on 3D reconstruction. Our method yields more accurate and robust geometric structures across diverse scenarios when compared to existing feed-forward approaches.}
    \label{fig:qualitative_comparison}
\end{figure}

Qualitative comparisons of point cloud reconstruction are shown in Fig.~\ref{fig:qualitative_comparison}. The results cover diverse scenarios, including outdoor scenes, indoor environments, and object-centric inputs. Compared with existing feed-forward approaches, ViGeo recovers cleaner and more complete 3D structures. Pi3~\cite{pi3} often introduces structural noise and checkerboard artifacts, while VGGT~\cite{vggt} tends to produce incomplete or fragmented geometry. In contrast, ViGeo better preserves object shapes, scene layouts, and spatial consistency, leading to more coherent 3D reconstructions.

\begin{figure}[htbp]
    \centering
    \includegraphics[width=\linewidth]{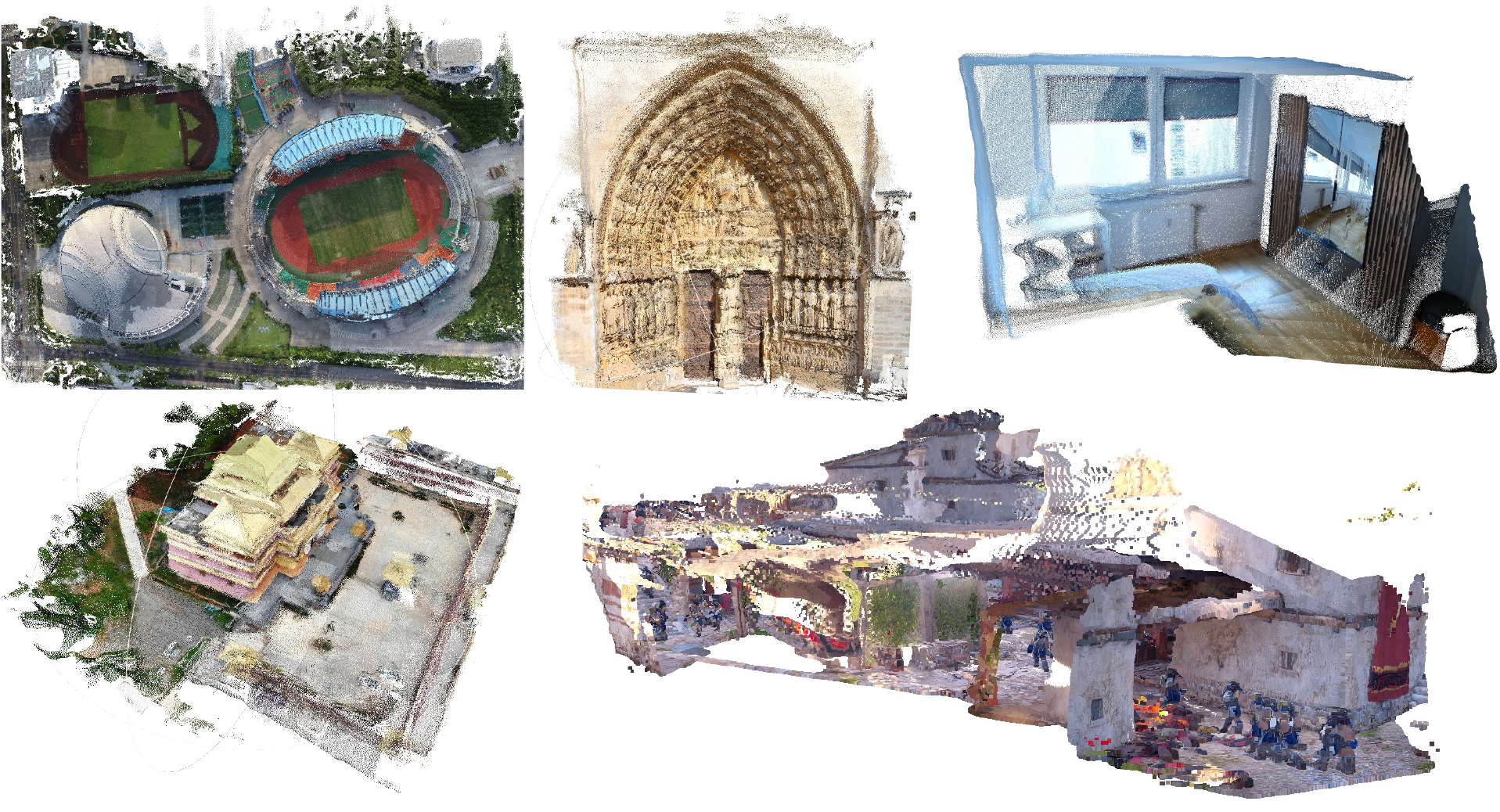}
    \caption{Additional point cloud visualizations. ViGeo produces accurate and realistic reconstructions with coherent geometry and fine structural details.}
    \label{fig:additional_point_cloud}
\end{figure}

Fig.~\ref{fig:additional_point_cloud} presents additional qualitative results of point cloud reconstruction. ViGeo produces accurate and realistic 3D structures with coherent global geometry and fine local details. The reconstructed point clouds preserve object boundaries and scene layouts, demonstrating the effectiveness of ViGeo in recovering dense and spatially consistent geometry from videos.

\subsubsection{Video Depth Estimation}
\begin{figure}[t]
    \centering
    \includegraphics[width=0.95\linewidth]{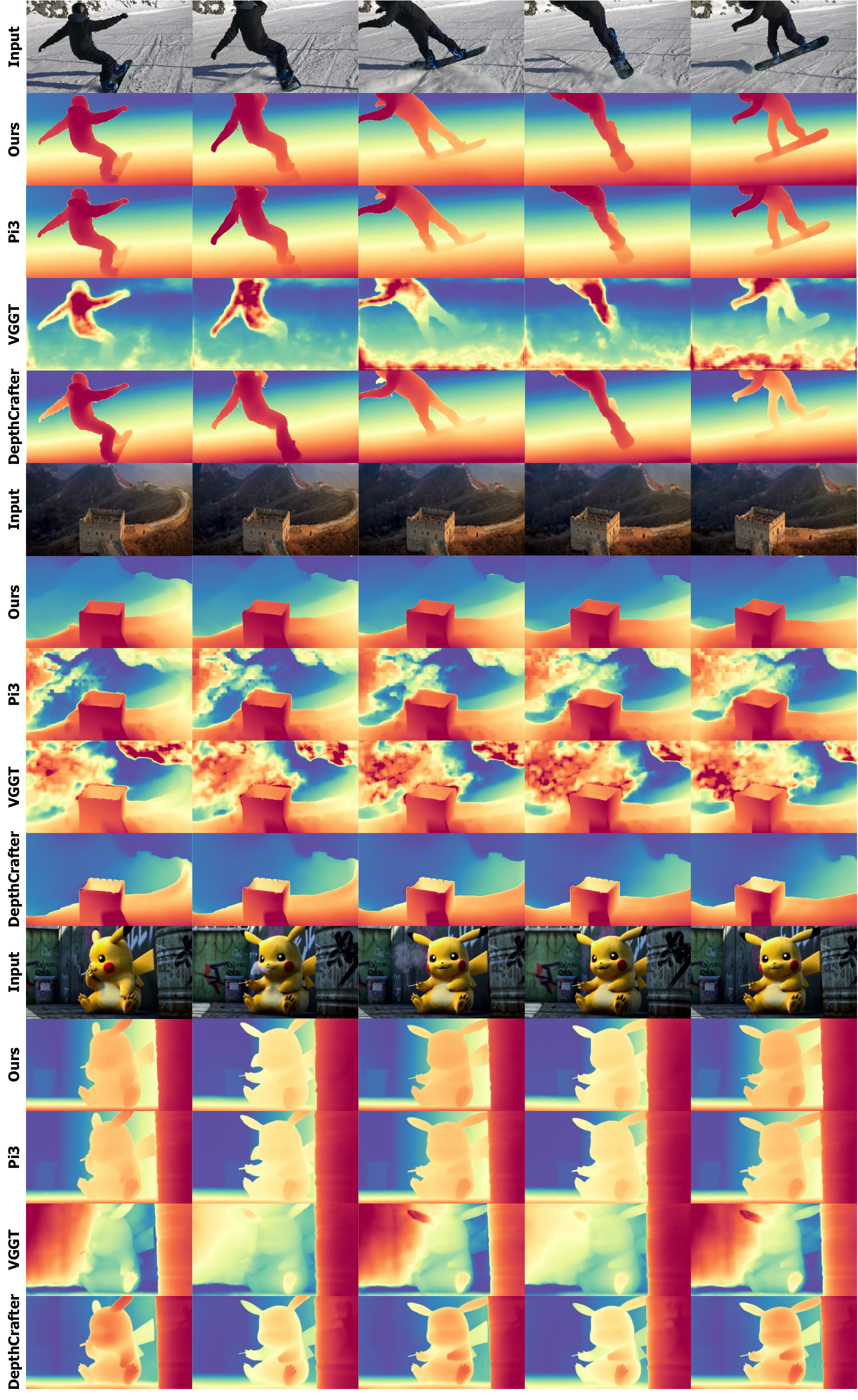}
    \caption{Qualitative results for video depth estimation. Compared with existing methods, ViGeo produces sharper, more accurate, and temporally stable depth.}
    \label{fig:video_compare}
\end{figure}

Fig.~\ref{fig:video_compare} compares video depth estimation results across consecutive frames. Beyond per-frame depth quality, our method maintains stable depth structures over time, especially for moving objects and camera-induced scene changes. VGGT and Pi3 often produce noisy responses and noticeable frame-to-frame depth fluctuations, causing unstable object shapes and inconsistent background geometry. DepthCrafter generates smoother results, but it tends to over-smooth large background regions and exhibits temporal inconsistency around dynamic objects. In contrast, our method preserves coherent depth ordering and stable object/background structures across frames, producing sharper and temporally consistent video depth predictions.

\subsubsection{Monocular Depth Estimation}
\begin{figure}[t]
    \centering
    \includegraphics[width=\linewidth]{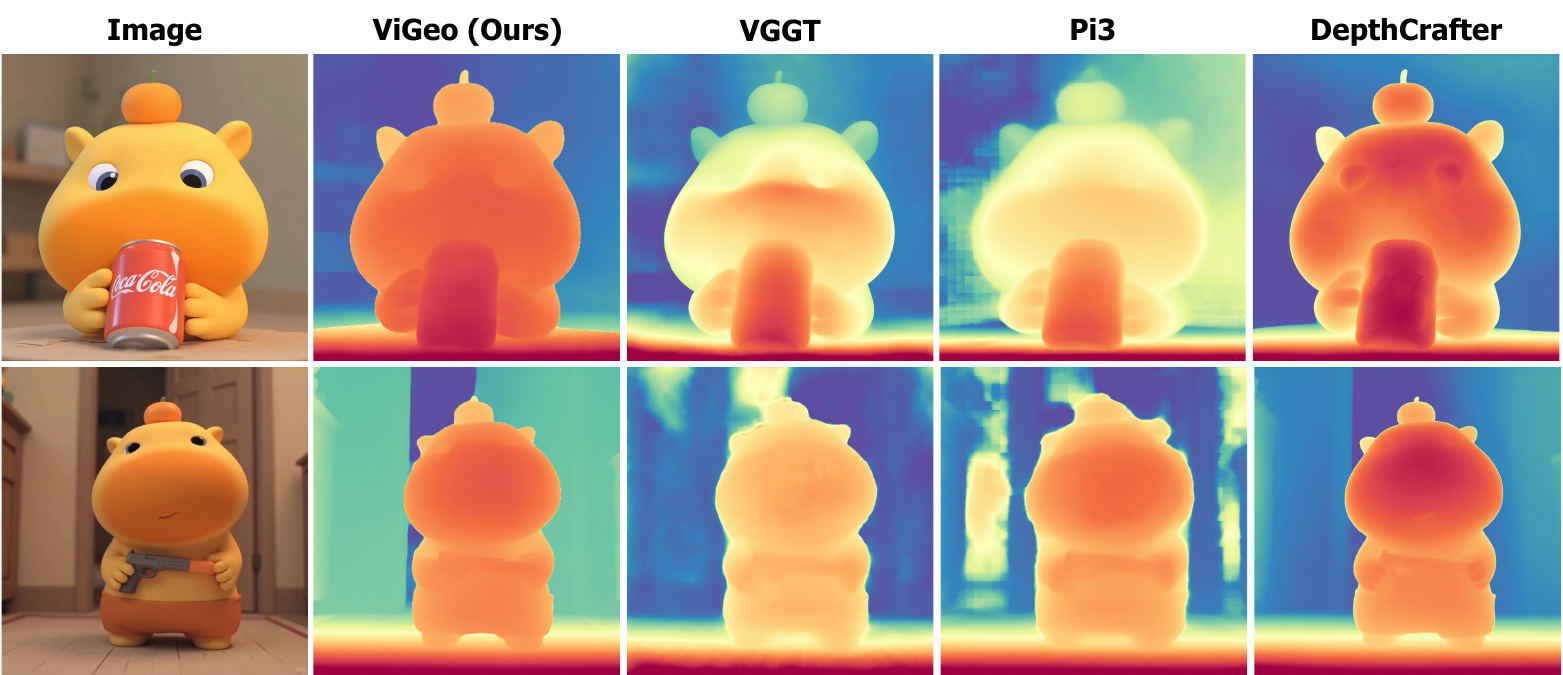}
    \caption{Qualitative results for monocular depth estimation. Compared with existing methods, ViGeo produces sharper depth boundaries and more accurate relative geometry.}
    \label{fig:mono_compare}
\end{figure}

Fig.~\ref{fig:mono_compare} compares monocular depth estimation results from a single input image. Our method produces sharp object boundaries and consistent relative geometry. Compared with VGGT and Pi3, our results better preserve clean foreground--background separation with fewer blurred or noisy structures. Although DepthCrafter yields sharp object contours, its background depth is overly smooth and its relative geometry is often inaccurate, leading to incorrect depth ordering and distorted scene structure. In contrast, our method maintains both crisp boundaries and geometrically plausible depth.

\subsubsection{Data Refinement}

\begin{figure}[t]
    \centering
    \includegraphics[width=\linewidth]{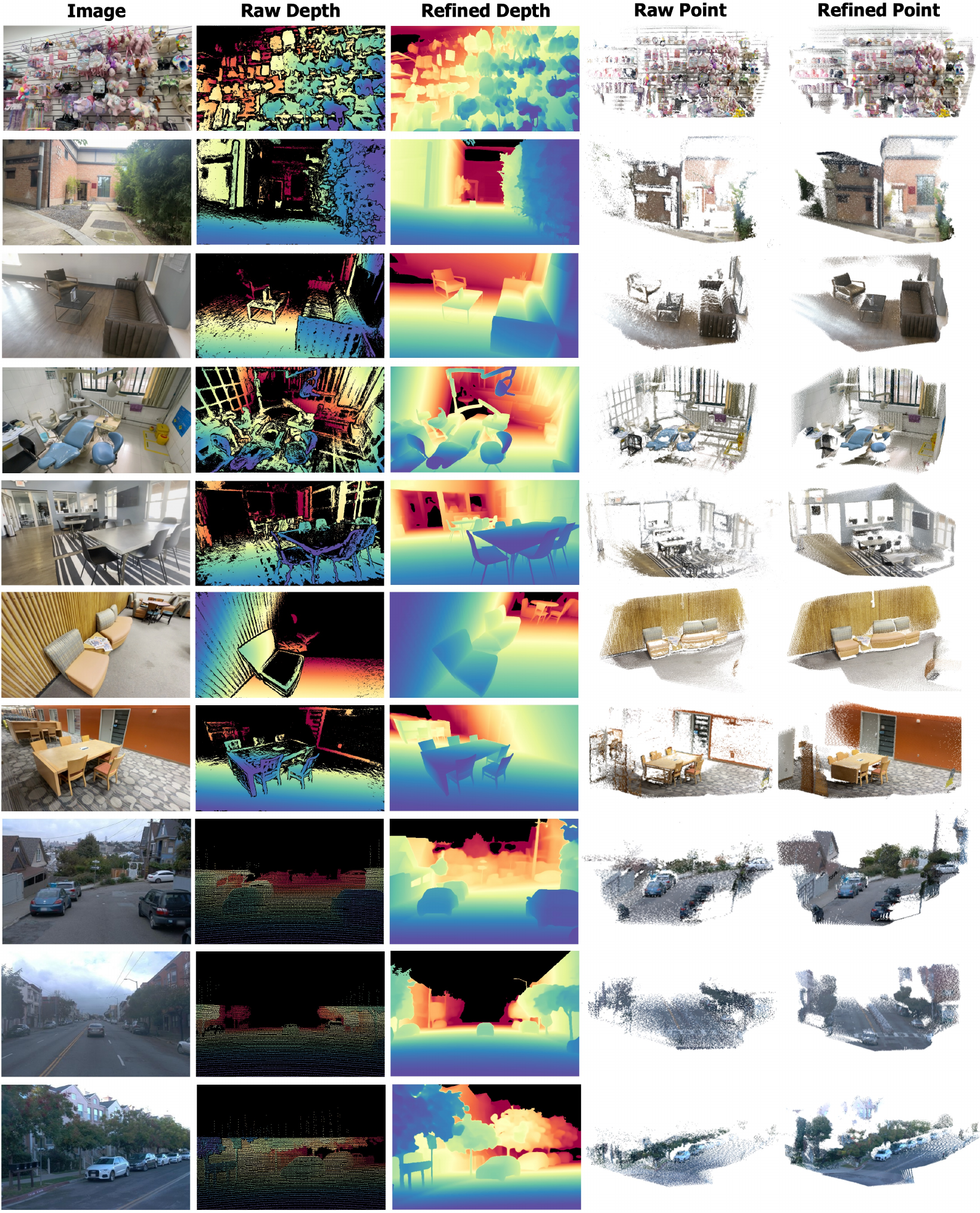}
    \caption{Qualitative results of the data refinement pipeline. We visualize the sparse raw measurements (cols. 2 \& 4) alongside our refined pseudo-labels (cols. 3 \& 5). While the raw depth suffers from extensive missing regions, our pipeline produces spatially dense and geometrically consistent labels. Notably, the refined point clouds recover solid surfaces and sharp structural details that are fragmented or entirely absent in the raw sensor outputs.}
    \label{fig:refine_demo}
\end{figure}

Fig.~\ref{fig:refine_demo} presents additional qualitative results of completion-based data refinement. Compared with raw depth maps that suffer from large missing regions, our pipeline produces dense and geometrically consistent labels. The refined point clouds further recover complete surfaces and sharp structures that are fragmented or absent in the raw sensor measurements.

\section{Discussion}
\label{sec:discussion}

\subsection{Limitations}
\label{app:limitations}

While ViGeo achieves promising results, high-resolution and 4D video geometry estimation remain challenging. High-resolution inputs may introduce additional computational cost, particularly for long sequences. Moreover, more explicit 4D representations could be explored to further improve temporal consistency in dynamic scenes. Despite these remaining challenges, we believe ViGeo provides a useful step toward scalable video geometry understanding.

\subsection{Broader Impacts}
\label{app:broader_impacts}
ViGeo provides a unified paradigm for video geometry estimation by supporting streaming, full-sequence, and long-video inference within a single feed-forward model. This may benefit applications that require dense and temporally consistent geometry, such as robotic perception, autonomous navigation, AR/VR, video editing, and 3D scene understanding. Beyond the model itself, our completion-based data refinement framework can serve as a reusable data engine for converting sparse and noisy real-world annotations into higher-quality geometric supervision, potentially facilitating the construction of larger and more reliable video geometry datasets.

Potential risks include privacy concerns when reconstructing real-world scenes from videos, as well as possible misuse in unauthorized mapping or surveillance. Since ViGeo is trained on public datasets, it may also inherit dataset biases and perform less reliably in underrepresented environments. For safety-critical applications, the model should be carefully validated under the target deployment conditions and used with appropriate safeguards.

\section{Conclusion}
In this paper, we present \textbf{ViGeo}, a feed-forward geometry foundation model for temporally consistent depth and surface normal estimation. By introducing \textbf{dynamic chunking attention}, ViGeo unifies streaming and full-sequence inference within a single transformer model. To improve supervision, we further develop a \textbf{robust data refinement framework} that converts sparse and noisy annotations into dense, coherent, and geometrically reliable targets. Experiments across multiple benchmarks show that ViGeo achieves state-of-the-art performance while maintaining strong spatial sharpness and temporal consistency. These results suggest that ViGeo provides a practical foundation for scalable video geometry estimation.

\clearpage
\bibliography{references}

@inproceedings{hypersim,
	title={Hypersim: A photorealistic synthetic dataset for holistic indoor scene understanding},
	author={Roberts, Mike and Ramapuram, Jason and Ranjan, Anurag and Kumar, Atulit and Bautista, Miguel Angel and Paczan, Nathan and Webb, Russ and Susskind, Joshua M},
	booktitle={Proceedings of the IEEE/CVF International Conference on Computer Vision},
	pages={10912--10922},
	year={2021}
}

@misc{lightwheelocc,
  title={LightwheelOcc: A 3D Occupancy Synthetic Dataset in Autonomous Driving} ,
  author={LightwheelAI and LightwheelOcc contributors},
  howpublished={\url{https://github.com/OpenDriveLab/LightwheelOcc}},
  year={2024}
}

@inproceedings{tartanair,
	title={Tartanair: A dataset to push the limits of visual slam},
	author={Wang, Wenshan and Zhu, Delong and Wang, Xiangwei and Hu, Yaoyu and Qiu, Yuheng and Wang, Chen and Hu, Yafei and Kapoor, Ashish and Scherer, Sebastian},
	booktitle={2020 IEEE/RSJ International Conference on Intelligent Robots and Systems},
	pages={4909--4916},
	year={2020},
}

@article{gtasfm,
  author       = {Kaixuan Wang and Shaojie Shen},
  title        = {Flow-Motion and Depth Network for Monocular Stereo and Beyond},
  journal      = {arXiv preprint arXiv:1909.05452},
  year        = {2019}
}

@inproceedings{pointodyssey,
	title={Pointodyssey: A large-scale synthetic dataset for long-term point tracking},
	author={Zheng, Yang and Harley, Adam W and Shen, Bokui and Wetzstein, Gordon and Guibas, Leonidas J},
	booktitle={Proceedings of the IEEE/CVF International Conference on Computer Vision},
	pages={19855--19865},
	year={2023}
}

@inproceedings{bedlam,
  title = {{BEDLAM}: A Synthetic Dataset of Bodies Exhibiting Detailed Lifelike Animated Motion},
  author = {Black, Michael J. and Patel, Priyanka and Tesch, Joachim and Yang, Jinlong}, 
  booktitle={Proceedings of the IEEE/CVF International Conference on Computer Vision},
  pages = {8726-8737},
  year = {2023}
}

@inproceedings{dynamicreplica,
  title={DynamicStereo: Consistent Dynamic Depth from Stereo Videos},
  author={Nikita Karaev and Ignacio Rocco and Benjamin Graham and Natalia Neverova and Andrea Vedaldi and Christian Rupprecht},
  booktitle={Proceedings of the IEEE/CVF International Conference on Computer Vision},
  year={2023}
}

@inproceedings{matrixcity,
	title={Matrixcity: A large-scale city dataset for city-scale neural rendering and beyond},
	author={Li, Yixuan and Jiang, Lihan and Xu, Linning and Xiangli, Yuanbo and Wang, Zhenzhi and Lin, Dahua and Dai, Bo},
	booktitle={Proceedings of the IEEE/CVF International Conference on Computer Vision},
	pages={3205--3215},
	year={2023}
}

@inproceedings{mvssynth,
	title={Deepmvs: Learning multi-view stereopsis},
	author={Huang, Po-Han and Matzen, Kevin and Kopf, Johannes and Ahuja, Narendra and Huang, Jia-Bin},
	booktitle={Proceedings of the IEEE/CVF conference on computer vision and pattern recognition},
	pages={2821--2830},
	year={2018}
}

@article{omniworld,
      title={OmniWorld: A Multi-Domain and Multi-Modal Dataset for 4D World Modeling}, 
      author={Yang Zhou and Yifan Wang and Jianjun Zhou and Wenzheng Chang and Haoyu Guo and Zizun Li and Kaijing Ma and Xinyue Li and Yating Wang and Haoyi Zhu and Mingyu Liu and Dingning Liu and Jiange Yang and Zhoujie Fu and Junyi Chen and Chunhua Shen and Jiangmiao Pang and Kaipeng Zhang and Tong He},
      journal={arXiv preprint arXiv:2509.12201},
      year={2025}
}

@inproceedings{synthia,
	title={The synthia dataset: A large collection of synthetic images for semantic segmentation of urban scenes},
	author={Ros, German and Sellart, Laura and Materzynska, Joanna and Vazquez, David and Lopez, Antonio M},
	booktitle={Proceedings of the IEEE/CVF conference on computer vision and pattern recognition},
	pages={3234--3243},
	year={2016}
}

@article{lotus,
  title={Lotus: Diffusion-based visual foundation model for high-quality dense prediction},
  author={He, Jing and Li, Haodong and Yin, Wei and Liang, Yixun and Li, Leheng and Zhou, Kaiqiang and Zhang, Hongbo and Liu, Bingbing and Chen, Ying-Cong},
  journal={arXiv preprint arXiv:2409.18124},
  year={2024}
}

@inproceedings{geowizard,
  title={Geowizard: Unleashing the diffusion priors for 3d geometry estimation from a single image},
  author={Fu, Xiao and Yin, Wei and Hu, Mu and Wang, Kaixuan and Ma, Yuexin and Tan, Ping and Shen, Shaojie and Lin, Dahua and Long, Xiaoxiao},
  booktitle={European Conference on Computer Vision},
  pages={241--258},
  year={2024},
  organization={Springer}
}

@inproceedings{omniobject3d,
    author = {Tong Wu and Jiarui Zhang and Xiao Fu and Yuxin Wang and Jiawei Ren and Liang Pan and Wayne Wu and Lei Yang and Jiaqi Wang and Chen Qian and Dahua Lin and Ziwei Liu},
    title = {OmniObject3D: Large-Vocabulary 3D Object Dataset for Realistic Perception, 
    Reconstruction and Generation},
    booktitle={Proceedings of the IEEE/CVF conference on computer vision and pattern recognition},
    year={2023}
}

@inproceedings{ase,
  author    = {Pan, Xiaqing and Charron, Nicholas and Yang, Yongqian and Peters, Scott and Whelan, Thomas and Kong, Chen and Parkhi, Omkar and Newcombe, Richard and Ren, Yuheng (Carl)},
  title     = {Aria Digital Twin: A New Benchmark Dataset for Egocentric 3D Machine Perception},
  booktitle = {Proceedings of the IEEE/CVF International Conference on Computer Vision},
  year      = {2023},
  pages     = {20133-20143}
}

@InProceedings{spring,
    author    = {Lukas Mehl and Jenny Schmalfuss and Azin Jahedi and Yaroslava Nalivayko and Andr\'es Bruhn},
    title     = {Spring: A High-Resolution High-Detail Dataset and Benchmark for Scene Flow, Optical Flow and Stereo},
    booktitle = {Proceedings of the IEEE/CVF conference on computer vision and pattern recognition},
    year      = {2023}
}

@InProceedings{wildrgb,
    author    = {Xia, Hongchi and Fu, Yang and Liu, Sifei and Wang, Xiaolong},
    title     = {RGBD Objects in the Wild: Scaling Real-World 3D Object Learning from RGB-D Videos},
    booktitle = {Proceedings of the IEEE/CVF conference on computer vision and pattern recognition},
    year      = {2024},
    pages     = {22378-22389}
}

@inproceedings{waymo, 
	title = {Scalability in Perception for Autonomous Driving: Waymo Open Dataset}, 
	author = {Sun, Pei and Kretzschmar, Henrik and Dotiwalla, Xerxes and Chouard, Aurelien and Patnaik, Vijaysai and Tsui, Paul and Guo, James and Zhou, Yin and Chai, Yuning and Caine, Benjamin and Vasudevan, Vijay and Han, Wei and Ngiam, Jiquan and Zhao, Hang and Timofeev, Aleksei and Ettinger, Scott and Krivokon, Maxim and Gao, Amy and Joshi, Aditya and Zhang, Yu and Shlens, Jonathon and Chen, Zhifeng and Anguelov, Dragomir}, 
	booktitle = {Proceedings of the IEEE/CVF conference on computer vision and pattern recognition}, 
	year = {2020} 
}

@inproceedings{arkitscenes,
    title={{ARK}itScenes - A Diverse Real-World Dataset for 3D Indoor Scene Understanding Using Mobile {RGB}-D Data},
    author={Gilad Baruch and Zhuoyuan Chen and Afshin Dehghan and Tal Dimry and Yuri Feigin and Peter Fu and Thomas Gebauer and Brandon Joffe and Daniel Kurz and Arik Schwartz and Elad Shulman},
    booktitle={Advances in Neural Information Processing Systems Datasets and Benchmarks Track},
    year={2021},
}

@inproceedings{scannetpp,
  title={ScanNet++: A High-Fidelity Dataset of 3D Indoor Scenes},
  author={Yeshwanth, Chandan and Liu, Yueh-Cheng and Nie{\ss}ner, Matthias and Dai, Angela},
  booktitle = {Proceedings of the IEEE/CVF International Conference on Computer Vision},
  year={2023}
}

@inproceedings{dl3dv,
  title     = {Dl3dv-10k: A large-scale scene dataset for deep learning-based 3d vision},
  author    = {Ling, Lu and Sheng, Yichen and Tu, Zhi and Zhao, Wentian and Xin, Cheng and Wan, Kun and Yu, Lantao and Guo, Qianyu and Yu, Zixun and Lu, Yawen and others},
  booktitle = {Proceedings of the IEEE/CVF conference on computer vision and pattern recognition},
  pages     = {22160--22169},
  year      = {2024}
}

@inproceedings{Blendedmvs,
  title     = {Blendedmvs: A large-scale dataset for generalized multi-view stereo networks},
  author    = {Yao, Yao and Luo, Zixin and Li, Shiwei and Zhang, Jingyang and Ren, Yufan and Zhou, Lei and Fang, Tian and Quan, Long},
  booktitle = {Proceedings of the IEEE/CVF conference on computer vision and pattern recognition},
  pages     = {1790--1799},
  year      = {2020}
}

@article{transphy3d,
  title={Diffusion Knows Transparency: Repurposing Video Diffusion for Transparent Object Depth and Normal Estimation}, 
  author={Shaocong Xu and Songlin Wei and Qizhe Wei and Zheng Geng and Hong Li and Licheng Shen and Qianpu Sun and Shu Han and Bin Ma and Bohan Li and Chongjie Ye and Yuhang Zheng and Nan Wang and Saining Zhang and Hao Zhao},
  journal={arXiv preprint arXiv:2512.23705},
  year={2025},
}

@inproceedings{adabins,
  title={Adabins: Depth estimation using adaptive bins},
  author={Bhat, Shariq Farooq and Alhashim, Ibraheem and Wonka, Peter},
  booktitle={Proceedings of the IEEE/CVF conference on computer vision and pattern recognition},
  pages={4009--4018},
  year={2021}
}

@article{eigendepth,
  title={Depth map prediction from a single image using a multi-scale deep network},
  author={Eigen, David and Puhrsch, Christian and Fergus, Rob},
  journal={Advances in Neural Information Processing Systems},
  volume={27},
  year={2014}
}

@inproceedings{vnl_mono,
  title={Enforcing geometric constraints of virtual normal for depth prediction},
  author={Yin, Wei and Liu, Yifan and Shen, Chunhua and Yan, Youliang},
  booktitle={Proceedings of the IEEE/CVF International Conference on Computer Vision},
  pages={5684--5693},
  year={2019}
}

@inproceedings{dorn,
  title={Deep ordinal regression network for monocular depth estimation},
  author={Fu, Huan and Gong, Mingming and Wang, Chaohui and Batmanghelich, Kayhan and Tao, Dacheng},
  booktitle={Proceedings of the IEEE/CVF conference on computer vision and pattern recognition},
  pages={2002--2011},
  year={2018}
}

@inproceedings{crfdepth,
  title={Neural window fully-connected crfs for monocular depth estimation},
  author={Yuan, Weihao and Gu, Xiaodong and Dai, Zuozhuo and Zhu, Siyu and Tan, Ping},
  booktitle={Proceedings of the IEEE/CVF conference on computer vision and pattern recognition},
  pages={3916--3925},
  year={2022}
}

@article{midas,
  title={Towards robust monocular depth estimation: Mixing datasets for zero-shot cross-dataset transfer},
  author={Ranftl, Ren{\'e} and Lasinger, Katrin and Hafner, David and Schindler, Konrad and Koltun, Vladlen},
  journal={IEEE transactions on pattern analysis and machine intelligence},
  volume={44},
  number={3},
  pages={1623--1637},
  year={2020}
}

@article{midasv3,
  title={Midas v3. 1--a model zoo for robust monocular relative depth estimation},
  author={Birkl, Reiner and Wofk, Diana and M{\"u}ller, Matthias},
  journal={arXiv preprint arXiv:2307.14460},
  year={2023}
}

@inproceedings{dpt,
  title={Vision transformers for dense prediction},
  author={Ranftl, Ren{\'e} and Bochkovskiy, Alexey and Koltun, Vladlen},
  booktitle={Proceedings of the IEEE/CVF International Conference on Computer Vision},
  pages={12179--12188},
  year={2021}
}

@article{depth_anything_v2,
  title={Depth Anything V2},
  author={Yang, Lihe and Kang, Bingyi and Huang, Zilong and Zhao, Zhen and Xu, Xiaogang and Feng, Jiashi and Zhao, Hengshuang},
  journal={arXiv:2406.09414},
  year={2024}
}

@inproceedings{depth_anything_v1,
  title={Depth Anything: Unleashing the Power of Large-Scale Unlabeled Data}, 
  author={Yang, Lihe and Kang, Bingyi and Huang, Zilong and Xu, Xiaogang and Feng, Jiashi and Zhao, Hengshuang},
  booktitle={Proceedings of the IEEE/CVF conference on computer vision and pattern recognition},
  year={2024}
}

@inproceedings{marigold,
  title={Repurposing diffusion-based image generators for monocular depth estimation},
  author={Ke, Bingxin and Obukhov, Anton and Huang, Shengyu and Metzger, Nando and Daudt, Rodrigo Caye and Schindler, Konrad},
  booktitle={Proceedings of the IEEE/CVF conference on computer vision and pattern recognition},
  pages={9492--9502},
  year={2024}
}

@inproceedings{stablediffusion,
  title={High-resolution image synthesis with latent diffusion models},
  author={Rombach, Robin and Blattmann, Andreas and Lorenz, Dominik and Esser, Patrick and Ommer, Bj{\"o}rn},
  booktitle={Proceedings of the IEEE/CVF conference on computer vision and pattern recognition},
  pages={10684--10695},
  year={2022}
}

@inproceedings{zerodepth,
  title={Towards zero-shot scale-aware monocular depth estimation},
  author={Guizilini, Vitor and Vasiljevic, Igor and Chen, Dian and Ambruș, Rareș and Gaidon, Adrien},
  booktitle={Proceedings of the IEEE/CVF International Conference on Computer Vision},
  pages={9233--9243},
  year={2023}
}

@inproceedings{metric3d,
  title={Metric3d: Towards zero-shot metric 3d prediction from a single image},
  author={Yin, Wei and Zhang, Chi and Chen, Hao and Cai, Zhipeng and Yu, Gang and Wang, Kaixuan and Chen, Xiaozhi and Shen, Chunhua},
  booktitle={Proceedings of the IEEE/CVF International Conference on Computer Vision},
  pages={9043--9053},
  year={2023}
}

@article{metric3dv2,
  title={Metric3D v2: A Versatile Monocular Geometric Foundation Model for Zero-shot Metric Depth and Surface Normal Estimation},
  author={Hu, Mu and Yin, Wei and Zhang, Chi and Cai, Zhipeng and Long, Xiaoxiao and Chen, Hao and Wang, Kaixuan and Yu, Gang and Shen, Chunhua and Shen, Shaojie},
  journal={arXiv preprint arXiv:2404.15506},
  year={2024}
}

@inproceedings{unidepth,
  title={UniDepth: Universal Monocular Metric Depth Estimation},
  author={Piccinelli, Luigi and Yang, Yung-Hsu and Sakaridis, Christos and Segu, Mattia and Li, Siyuan and Van Gool, Luc and Yu, Fisher},
  booktitle={Proceedings of the IEEE/CVF conference on computer vision and pattern recognition},
  pages={10106--10116},
  year={2024}
}

@article{unidepthv2,
  title={UniDepthV2: Universal Monocular Metric Depth Estimation Made Simpler},
  author={Piccinelli, Luigi and Sakaridis, Christos and Yang, Yung-Hsu and Segu, Mattia and Li, Siyuan and Abbeloos, Wim and Van Gool, Luc},
  journal={arXiv preprint arXiv:2502.20110},
  year={2025}
}

@article{dinov2,
  title={Dinov2: Learning robust visual features without supervision},
  author={Oquab, Maxime and Darcet, Timoth{\'e}e and Moutakanni, Th{\'e}o and Vo, Huy and Szafraniec, Marc and Khalidov, Vasil and Fernandez, Pierre and Haziza, Daniel and Massa, Francisco and El-Nouby, Alaaeldin and others},
  journal={arXiv preprint arXiv:2304.07193},
  year={2023}
}

@article{fe2e,
  title={From Editor to Dense Geometry Estimator},
  author={Wang, JiYuan and Lin, Chunyu and Sun, Lei and Liu, Rongying and Nie, Lang and Li, Mingxing and Liao, Kang and Chu, Xiangxiang and Zhao, Yao},
  journal={arXiv preprint arXiv:2509.04338},
  year={2025}
}

@inproceedings{colmap,
  title={Structure-from-motion revisited},
  author={Schonberger, Johannes L and Frahm, Jan-Michael},
  booktitle={Proceedings of the IEEE/CVF conference on computer vision and pattern recognition},
  pages={4104--4113},
  year={2016}
}

@inproceedings{dust3r,
  author       = {Shuzhe Wang and Vincent Leroy and Yohann Cabon and Boris Chidlovskii and J{\'{e}}r{\^{o}}me Revaud},
  title        = {DUSt3R: Geometric 3D Vision Made Easy},
  booktitle    = {{Proceedings of the IEEE/CVF conference on computer vision and pattern recognition}},
  pages        = {20697--20709},
  year         = {2024}
}

@InProceedings{moge,
    author    = {Wang, Ruicheng and Xu, Sicheng and Dai, Cassie and Xiang, Jianfeng and Deng, Yu and Tong, Xin and Yang, Jiaolong},
    title     = {MoGe: Unlocking Accurate Monocular Geometry Estimation for Open-Domain Images with Optimal Training Supervision},
    booktitle = {Proceedings of the IEEE/CVF conference on computer vision and pattern recognition},
    year      = {2025},
    pages     = {5261-5271}
}

@inproceedings{da3,
title={Depth Anything 3: Recovering the Visual Space from Any Views},
author={Haotong Lin and Sili Chen and Jun Hao Liew and Donny Y. Chen and Zhenyu Li and Yang Zhao and Sida Peng and Hengkai Guo and Xiaowei Zhou and Guang Shi and Jiashi Feng and Bingyi Kang},
booktitle={International Conference on Learning Representations},
year={2026},
}

@inproceedings{moge2,
title={MoGe-2: Accurate Monocular Geometry with Metric Scale and Sharp Details},
author={Ruicheng Wang and Sicheng Xu and Yue Dong and Yu Deng and Jianfeng Xiang and Zelong Lv and Guangzhong Sun and Xin Tong and Jiaolong Yang},
booktitle={Advances in Neural Information Processing Systems},
year={2025}
}

@article{dens3r,
      title={Dens3R: A Foundation Model for 3D Geometry Prediction}, 
      author={Xianze Fang and Jingnan Gao and Zhe Wang and Zhuo Chen and Xingyu Ren and Jiangjing Lyu and Qiaomu Ren and Zhonglei Yang and Xiaokang Yang and Yichao Yan and Chengfei Lyu},
      journal={arXiv preprint arXiv:2507.16290},
      year={2025}
}

@article{more,
  title={MoRE: 3D Visual Geometry Reconstruction Meets Mixture-of-Experts}, 
  author={Jingnan Gao and Zhe Wang and Xianze Fang and Xingyu Ren and Zhuo Chen and Shengqi Liu and Yuhao Cheng and Jiangjing Lyu and Xiaokang Yang and Yichao Yan},
  journal={arXiv preprint arXiv:2510.27234},
  year={2025}
}

@inproceedings{ldcm,
title={Large Depth Completion Model from Sparse Observations},
author={Zhu Yu and zhengyi zhao and Runmin Zhang and Lingteng Qiu and Si-Yuan Cao and Kejie Qiu and Yisheng He and Siyu Zhu and Zilong Dong and Hui-liang Shen},
booktitle={International Conference on Learning Representations},
year={2026}
}

@inproceedings{vggt,
  title={Vggt: Visual geometry grounded transformer},
  author={Wang, Jianyuan and Chen, Minghao and Karaev, Nikita and Vedaldi, Andrea and Rupprecht, Christian and Novotny, David},
  booktitle={Proceedings of the IEEE/CVF conference on computer vision and pattern recognition},
  pages={5294--5306},
  year={2025}
}

@inproceedings{mapanything,
title={MapAnything: Universal Feed-Forward Metric 3D Reconstruction},
author={Nikhil Varma Keetha and Norman M{\"u}ller and Johannes Sch{\"o}nberger and Lorenzo Porzi and Yuchen Zhang and Tobias Fischer and Arno Knapitsch and Duncan Zauss and Ethan Weber and Nelson Antunes and Jonathon Luiten and Manuel Lopez-Antequera and Samuel Rota Bul{\`o} and Christian Richardt and Deva Ramanan and Sebastian Scherer and Peter Kontschieder},
booktitle={3DV},
year={2026},
}

@inproceedings{pi3,
title={{$\pi^3$}: Permutation-Equivariant Visual Geometry Learning},
author={Yifan Wang and Jianjun Zhou and Haoyi Zhu and Wenzheng Chang and Yang Zhou and Zizun Li and Junyi Chen and Jiangmiao Pang and Chunhua Shen and Tong He},
booktitle={International Conference on Learning Representations},
year={2026}
}

@inproceedings{fast3r,
  title={Fast3r: Towards 3d reconstruction of 1000+ images in one forward pass},
  author={Yang, Jianing and Sax, Alexander and Liang, Kevin J and Henaff, Mikael and Tang, Hao and Cao, Ang and Chai, Joyce and Meier, Franziska and Feiszli, Matt},
  booktitle={Proceedings of the IEEE/CVF conference on computer vision and pattern recognition},
  pages={21924--21935},
  year={2025}
}

@inproceedings{mvdust3r,
  title={Mv-dust3r+: Single-stage scene reconstruction from sparse views in 2 seconds},
  author={Tang, Zhenggang and Fan, Yuchen and Wang, Dilin and Xu, Hongyu and Ranjan, Rakesh and Schwing, Alexander and Yan, Zhicheng},
  booktitle={Proceedings of the IEEE/CVF conference on computer vision and pattern recognition},
  pages={5283--5293},
  year={2025}
}

@inproceedings{monst3r,
  title={MonST3R: A Simple Approach for Estimating Geometry in the Presence of Motion},
  author={Zhang, Junyi and Herrmann, Charles and Hur, Junhwa and Jampani, Varun and Darrell, Trevor and Cole, Forrester and Sun, Deqing and Yang, Ming-Hsuan},
  booktitle={International Conference on Learning Representations},
  year={2025},
}

@inproceedings{mast3r,
  title={Grounding image matching in 3d with mast3r},
  author={Leroy, Vincent and Cabon, Yohann and Revaud, J{\'e}r{\^o}me},
  booktitle={European Conference on Computer Vision},
  pages={71--91},
  year={2024}
}

@article{luo2020consistent,
  title={Consistent video depth estimation},
  author={Luo, Xuan and Huang, Jia-Bin and Szeliski, Richard and Matzen, Kevin and Kopf, Johannes},
  journal={ACM Transactions on Graphics},
  volume={39},
  number={4},
  pages={71--1},
  year={2020}
}

@inproceedings{nvds,
  title={Neural video depth stabilizer},
  author={Wang, Yiran and Shi, Min and Li, Jiaqi and Huang, Zihao and Cao, Zhiguo and Zhang, Jianming and Xian, Ke and Lin, Guosheng},
  booktitle={Proceedings of the IEEE/CVF International Conference on Computer Vision},
  pages={9466--9476},
  year={2023}
}

@article{depthanyvideo,
  title={Depth Any Video with Scalable Synthetic Data},
  author={Yang, Honghui and Huang, Di and Yin, Wei and Shen, Chunhua and Liu, Haifeng and He, Xiaofei and Lin, Binbin and Ouyang, Wanli and He, Tong},
  journal={arXiv preprint arXiv:2410.10815},
  year={2024}
}

@article{chronodepth,
  title={Learning Temporally Consistent Video Depth from Video Diffusion Priors},
  author={Shao, Jiahao and Yang, Yuanbo and Zhou, Hongyu and Zhang, Youmin and Shen, Yujun and Poggi, Matteo and Liao, Yiyi},
  journal={arXiv preprint arXiv:2406.01493},
  year={2024}
}

@article{depthcrafter,
  title={Depthcrafter: Generating consistent long depth sequences for open-world videos},
  author={Hu, Wenbo and Gao, Xiangjun and Li, Xiaoyu and Zhao, Sijie and Cun, Xiaodong and Zhang, Yong and Quan, Long and Shan, Ying},
  journal={arXiv preprint arXiv:2409.02095},
  year={2024}
}

@InProceedings{vda,
    author    = {Chen, Sili and Guo, Hengkai and Zhu, Shengnan and Zhang, Feihu and Huang, Zilong and Feng, Jiashi and Kang, Bingyi},
    title     = {Video Depth Anything: Consistent Depth Estimation for Super-Long Videos},
    booktitle = {Proceedings of the IEEE/CVF conference on computer vision and pattern recognition},
    year      = {2025},
    pages     = {22831-22840}
}

@inproceedings{stream3r,
    title={{ST}ream3R: Scalable Sequential 3D Reconstruction with Causal Transformer},
    author={Yushi LAN and Yihang Luo and Fangzhou Hong and Shangchen Zhou and Honghua Chen and Zhaoyang Lyu and Bo Dai and Shuai Yang and Chen Change Loy and Xingang Pan},
    booktitle={International Conference on Learning Representations},
    year={2026}
  }

@article{streamvggt,
  title={Streaming {4D} Visual Geometry Transformer},
  author={Zhuo, Dong and Zheng, Wenzhao and Guo, Jiahe and Wu, Yuqi and Zhou, Jie and Lu, Jiwen},
  journal={arXiv preprint arXiv:2507.11539},
  year={2025}
}

@inproceedings{cut3r,
  title={Continuous 3d perception model with persistent state},
  author={Wang, Qianqian and Zhang, Yifei and Holynski, Aleksander and Efros, Alexei A and Kanazawa, Angjoo},
  booktitle={Proceedings of the IEEE/CVF conference on computer vision and pattern recognition},
  pages={10510--10522},
  year={2025}
}

@article{spann3r,
  title={3D Reconstruction with Spatial Memory},
  author={Wang, Hengyi and Agapito, Lourdes},
  journal={arXiv preprint arXiv:2408.16061},
  year={2024}
}

@inproceedings{dsine,
  title={Rethinking inductive biases for surface normal estimation},
  author={Bae, Gwangbin and Davison, Andrew J},
  booktitle={Proceedings of the IEEE/CVF conference on computer vision and pattern recognition},
  pages={9535--9545},
  year={2024}
}

@inproceedings{bae2021estimating,
  title={Estimating and exploiting the aleatoric uncertainty in surface normal estimation},
  author={Bae, Gwangbin and Budvytis, Ignas and Cipolla, Roberto},
  booktitle={Proceedings of the IEEE/CVF International Conference on Computer Vision},
  pages={13137--13146},
  year={2021}
}

@article{qi2020geonet,
  title={Geonet++: Iterative geometric neural network with edge-aware refinement for joint depth and surface normal estimation},
  author={Qi, Xiaojuan and Liu, Zhengzhe and Liao, Renjie and Torr, Philip HS and Urtasun, Raquel and Jia, Jiaya},
  journal={IEEE Transactions on Pattern Analysis and Machine Intelligence},
  volume={44},
  number={2},
  pages={969--984},
  year={2020}
}

@article{marigoldv2,
  title={Marigold: Affordable Adaptation of Diffusion-Based Image Generators for Image Analysis},
  author={Ke, B and Qu, K and Wang, T and Metzger, N and Huang, S and Li, B and Obukhov, A and Schindler, K},
  journal={arXiv preprint arXiv:2505.09358},
  year={2025}
}

@article{drioid,
  title={Droid-slam: Deep visual slam for monocular, stereo, and rgb-d cameras},
  author={Teed, Zachary and Deng, Jia},
  journal={Advances in neural information processing systems},
  volume={34},
  pages={16558--16569},
  year={2021}
}

@inproceedings{vggsfm,
  title={Vggsfm: Visual geometry grounded deep structure from motion},
  author={Wang, Jianyuan and Karaev, Nikita and Rupprecht, Christian and Novotny, David},
  booktitle={Proceedings of the IEEE/CVF conference on computer vision and pattern recognition},
  pages={21686--21697},
  year={2024}
}

@inproceedings{align3r,
  title={Align3r: Aligned monocular depth estimation for dynamic videos},
  author={Lu, Jiahao and Huang, Tianyu and Li, Peng and Dou, Zhiyang and Lin, Cheng and Cui, Zhiming and Dong, Zhen and Yeung, Sai-Kit and Wang, Wenping and Liu, Yuan},
  booktitle={Proceedings of the IEEE/CVF conference on computer vision and pattern recognition},
  pages={22820--22830},
  year={2025}
}

@article{svd,
  title={Stable video diffusion: Scaling latent video diffusion models to large datasets},
  author={Blattmann, Andreas and Dockhorn, Tim and Kulal, Sumith and Mendelevitch, Daniel and Kilian, Maciej and Lorenz, Dominik and Levi, Yam and English, Zion and Voleti, Vikram and Letts, Adam and others},
  journal={arXiv preprint arXiv:2311.15127},
  year={2023}
}

@article{vit,
  title={An image is worth 16x16 words: Transformers for image recognition at scale},
  author={Dosovitskiy, Alexey and Beyer, Lucas and Kolesnikov, Alexander and Weissenborn, Dirk and Zhai, Xiaohua and Unterthiner, Thomas and Dehghani, Mostafa and Minderer, Matthias and Heigold, Georg and Gelly, Sylvain and others},
  journal={arXiv preprint arXiv:2010.11929},
  year={2020}
}

@article{lingbodepth,
  title={Masked Depth Modeling for Spatial Perception},
  author={Tan, Bin and Sun, Changjiang and Qin, Xiage and Adai, Hanat and Fu, Zelin and Zhou, Tianxiang and Zhang, Han and Xu, Yinghao and Zhu, Xing and Shen, Yujun and others},
  journal={arXiv preprint arXiv:2601.17895},
  year={2026}
}

@inproceedings{flashdepth,
  title={Flashdepth: Real-time streaming video depth estimation at 2k resolution},
  author={Chou, Gene and Xian, Wenqi and Yang, Guandao and Abdelfattah, Mohamed and Hariharan, Bharath and Snavely, Noah and Yu, Ning and Debevec, Paul},
  booktitle={Proceedings of the IEEE/CVF International Conference on Computer Vision},
  pages={9638--9648},
  year={2025}
}

@article{carlaocc,
  title={An Instance-Centric Panoptic Occupancy Prediction Benchmark for Autonomous Driving},
  author={Feng, Yi and Guo, Zizhan and Ma, Yu and Wang, Hanli and Fan, Rui and others},
  journal={arXiv preprint arXiv:2603.27238},
  year={2026}
}

@inproceedings{tartanground,
  title={Tartanground: A large-scale dataset for ground robot perception and navigation},
  author={Patel, Manthan and Yang, Fan and Qiu, Yuheng and Cadena, Cesar and Scherer, Sebastian and Hutter, Marco and Wang, Wenshan},
  booktitle={2025 IEEE/RSJ International Conference on Intelligent Robots and Systems (IROS)},
  pages={20524--20531},
  year={2025},
  organization={IEEE}
}

@article{sintel,
  title={Transformerfusion: Monocular rgb scene reconstruction using transformers},
  author={Bozic, Aljaz and Palafox, Pablo and Thies, Justus and Dai, Angela and Nie{\ss}ner, Matthias},
  journal={Advances in Neural Information Processing Systems},
  volume={34},
  pages={1403--1414},
  year={2021}
}

@inproceedings{bonn,
  title={ReFusion: 3D reconstruction in dynamic environments for RGB-D cameras exploiting residuals},
  author={Palazzolo, Emanuele and Behley, Jens and Lottes, Philipp and Giguere, Philippe and Stachniss, Cyrill},
  booktitle={IEEE/RSJ International Conference on Intelligent Robots and Systems},
  pages={7855--7862},
  year={2019}
}

@inproceedings{sevenscenes,
  title={Scene Coordinate Regression Forests for Camera Relocalization in RGB-D Images},
  author={Shotton, Jamie and Glocker, Ben and Zach, Christopher and Izadi, Shahram and Criminisi, Antonio and Fitzgibbon, Andrew},
  booktitle={Proceedings of the IEEE Conference on Computer Vision and Pattern Recognition},
  pages={2930--2937},
  year={2013}
}

@inproceedings{neuralrgbd,
  title={Neural RGB-D Surface Reconstruction},
  author={Azinovi{\'c}, Dejan and Martin-Brualla, Ricardo and Goldman, Dan B and Nie{\ss}ner, Matthias and Thies, Justus},
  booktitle={Proceedings of the IEEE/CVF Conference on Computer Vision and Pattern Recognition},
  pages={6290--6301},
  year={2022}
}

@article{kitti,
  title={Vision meets robotics: The kitti dataset},
  author={Geiger, Andreas and Lenz, Philip and Stiller, Christoph and Urtasun, Raquel},
  journal={The International Journal of Robotics Research},
  volume={32},
  number={11},
  pages={1231--1237},
  year={2013},
}

@inproceedings{geometrycrafter,
  title={Geometrycrafter: Consistent geometry estimation for open-world videos with diffusion priors},
  author={Xu, Tian-Xing and Gao, Xiangjun and Hu, Wenbo and Li, Xiaoyu and Zhang, Song-Hai and Shan, Ying},
  booktitle={Proceedings of the IEEE/CVF International Conference on Computer Vision},
  pages={6632--6644},
  year={2025}
}

@inproceedings{hammer,
  title={On the importance of accurate geometry data for dense 3d vision tasks},
  author={Jung, HyunJun and Ruhkamp, Patrick and Zhai, Guangyao and Brasch, Nikolas and Li, Yitong and Verdie, Yannick and Song, Jifei and Zhou, Yiren and Armagan, Anil and Ilic, Slobodan and others},
  booktitle={Proceedings of the IEEE/CVF Conference on Computer Vision and Pattern Recognition},
  pages={780--791},
  year={2023}
}

@article{stablenormal,
  author       = {Chongjie Ye and
                  Lingteng Qiu and
                  Xiaodong Gu and
                  Qi Zuo and
                  Yushuang Wu and
                  Zilong Dong and
                  Liefeng Bo and
                  Yuliang Xiu and
                  Xiaoguang Han},
  title        = {StableNormal: Reducing Diffusion Variance for Stable and Sharp Normal},
  journal      = {{ACM} Trans. Graph.},
  volume       = {43},
  number       = {6},
  pages        = {250:1--250:18},
  year         = {2024}
}

@InProceedings{normalcrafter,
    author    = {Bin, Yanrui and Hu, Wenbo and Wang, Haoyuan and Chen, Xinya and Wang, Bing},
    title     = {NormalCrafter: Learning Temporally Consistent Normals from Video Diffusion Priors},
    booktitle = {Proceedings of the IEEE/CVF International Conference on Computer Vision},
    year      = {2025},
    pages     = {8330-8339}
}

@misc{infinitevggt,
        title={InfiniteVGGT: Visual Geometry Grounded Transformer for Endless Streams}, 
        author={Shuai Yuan and Yantai Yang and Xiaotian Yang and Xupeng Zhang and Zhonghao Zhao and Lingming Zhang and Zhipeng Zhang},
        journal={arXiv preprint arXiv:2601.02281},
        year={2026}
}

@inproceedings{nyuv2,
	title={Indoor segmentation and support inference from rgbd images},
	author={Silberman, Nathan and Hoiem, Derek and Kohli, Pushmeet and Fergus, Rob},
	booktitle={Proceedings of the European Conference on Computer Vision},
	pages={746--760},
	year={2012}
}

@article{geovideogen,
  title={Geometry-aware 4d video generation for robot manipulation},
  author={Liu, Zeyi and Li, Shuang and Cousineau, Eric and Feng, Siyuan and Burchfiel, Benjamin and Song, Shuran},
  journal={arXiv preprint arXiv:2507.01099},
  year={2025}
}

@article{depthar,
  title={Depth from motion for smartphone AR},
  author={Valentin, Julien and Kowdle, Adarsh and Barron, Jonathan T and Wadhwa, Neal and Dzitsiuk, Max and Schoenberg, Michael and Verma, Vivek and Csaszar, Ambrus and Turner, Eric and Dryanovski, Ivan and others},
  journal={ACM Transactions on Graphics},
  volume={37},
  number={6},
  pages={1--19},
  year={2018},
}

@article{depthnav,
  title={Enhanced depth navigation through augmented reality depth mapping in patients with low vision},
  author={Angelopoulos, Anastasios Nikolas and Ameri, Hossein and Mitra, Debbie and Humayun, Mark},
  journal={Scientific reports},
  volume={9},
  number={1},
  pages={11230},
  year={2019}
}

@inproceedings{depthedit,
  title={Pix2video: Video editing using image diffusion},
  author={Ceylan, Duygu and Huang, Chun-Hao P and Mitra, Niloy J},
  booktitle={Proceedings of the IEEE/CVF international conference on computer vision},
  pages={23206--23217},
  year={2023}
}

@inproceedings{pointdc,
  title={Aggregating feature point cloud for depth completion},
  author={Yu, Zhu and Sheng, Zehua and Zhou, Zili and Luo, Lun and Cao, Si-Yuan and Gu, Hong and Zhang, Huaqi and Shen, Hui-Liang},
  booktitle={Proceedings of the IEEE/CVF international conference on computer vision},
  pages={8732--8743},
  year={2023}
}
\bibliographystyle{ieee_fullname}
\end{document}